%% file: ViterbiNetFull_v07_Arxiv.tex
\newif\ifsingle
\singletrue 

\newif\ifproofs

\ifsingle
\documentclass[12pt,draftclsnofoot, onecolumn]{IEEEtran}		
\else		
\documentclass[10pt,final, twocolumn]{IEEEtran}
\fi

\usepackage{times}
\usepackage{amsmath,dsfont}
\usepackage{amssymb,amsthm}
\usepackage{epsfig,verbatim}
\usepackage{subfigure}
\usepackage{setspace}
\usepackage{color}
\usepackage{cite}
\usepackage{epstopdf}
\usepackage{graphics}
\usepackage{accents}
\usepackage{acronym}
\usepackage[bookmarks,colorlinks]{hyperref}
\usepackage{booktabs}
\usepackage{mathtools}
\usepackage{algorithm}
\usepackage{algorithmic}
\usepackage{enumitem}

\input{Preamble.tex}

\ifsingle
\newcommand{\figWidth}{0.65\columnwidth}
\newcommand{\figSpace}{\vspace{-0.2cm}}
\newcommand{\includefig}[1]{\includegraphics[width = 0.65\columnwidth]{#1} 	\vspace{-0.2cm}}
\else
\newcommand{\figWidth}{\columnwidth}
\newcommand{\includefig}[1]{\includegraphics[width = 0.9\columnwidth]{#1} 	\vspace{-0.2cm}}
\newcommand{\figSpace}{\vspace{-0.6cm}}
\fi 

\newcommand{\revision}[1]{#1}

\IEEEoverridecommandlockouts 

\title{ViterbiNet: A Deep Learning Based Viterbi Algorithm for Symbol Detection 
}
\author{
	\IEEEauthorblockN{Nir Shlezinger,  Nariman Farsad, Yonina C. Eldar, and Andrea J. Goldsmith\\
	} 
	\thanks{Parts of this work were accepted for presentation in the 2019 IEEE International Workshop on Signal Processing Advances in Wireless Communications (SPAWC), Cannes, France.}
	\thanks{This work was supported in part by the  US - Israel Binational Science Foundation	under grant No. 2026094,  by the Israel Science Foundation under grant No. 0100101, and by the Office of the Naval Research under grant No. 18-1-2191.
		}
	\thanks{
		N. Shlezinger  and Y. C. Eldar are with the Faculty of Math and CS, Weizmann Institute of Science, Rehovot, Israel (e-mail: nirshlezinger1@gmail.com; yonina@weizmann.ac.il). 	
}	
	\thanks{ 
		 N. Farsad  and A. J. Goldsmith are with the Department of EE, Stanford, Palo Alto, CA (e-mail:  nfarsad@stanford.edu; andrea@wsl.stanford.edu).  	
	}

	\vspace{-1.0cm}
	
}
\vspace{-0.75cm}

\begin{document}
	
	\maketitle
	\pagestyle{plain}
	\thispagestyle{plain}
	\begin{abstract}
	 Symbol detection plays an important role in the implementation of digital receivers. 
	 In this work, we propose ViterbiNet, which is a data-driven symbol detector that does not require channel state information (CSI). ViterbiNet is obtained by integrating deep neural networks (DNNs) into the Viterbi algorithm.  We identify the specific parts of the Viterbi algorithm that are channel-model-based, and design a DNN to implement only those computations, leaving the rest of the algorithm structure intact. 
	 We then propose a meta-learning based approach to train ViterbiNet online based on recent decisions, allowing the receiver to track dynamic channel conditions without requiring new training samples for every coherence block.
	  Our numerical evaluations demonstrate that the performance of ViterbiNet, which is ignorant of the CSI, approaches that of the  CSI-based Viterbi algorithm, and is capable of tracking time-varying channels without needing instantaneous CSI or additional training data. Moreover, unlike conventional Viterbi detection, ViterbiNet is robust to CSI uncertainty, and it can be reliably implemented in complex channel models with constrained computational burden. More broadly, our results demonstrate the conceptual benefit of designing communication systems to that integrate DNNs into established  algorithms.
	\end{abstract}

	\vspace{-0.4cm}
	\section{Introduction}
	\vspace{-0.1cm}
	A fundamental task of a digital receiver is to reliably recover the transmitted symbols from the observed channel output. This task is commonly referred to as {\em symbol detection}. Conventional symbol detection algorithms, such as those based on the \ac{map} rule, require complete knowledge of the underlying channel model and its parameters. 
	Consequently, in some cases, conventional methods cannot be applied, particularly when the channel model is highly complex, poorly understood, or does not well-capture the underlying physics of the system. 
	Furthermore, when the channel models are known, many detection algorithms rely on the instantaneous \ac{csi}, i.e., the instantaneous parameters of the channel model, for detection. Therefore, conventional channel-model-based techniques require the instantaneous \ac{csi} to be estimated. However, this  process entails overhead, which decreases the data transmission rate. Moreover, inaccurate  \ac{csi} estimation typically degrades the detection performance.

	One of the most common \ac{csi}-based symbol detection methods is the iterative scheme proposed by  Viterbi in \cite{Viterbi:67}, known as the Viterbi algorithm. The Viterbi algorithm is an efficient symbol detector that is capable of  achieving the minimal probability of error in recovering the transmitted symbols for channels obeying a Markovian input-output stochastic relationship, which is encountered in many practical channels \cite{Forney:73}. Since the Viterbi algorithm   requires the receiver to know the exact statistical relationship relating the channel input and output,   the receiver must have full instantaneous \ac{csi}.

	An alternative data-driven approach is based on \ac{ml}. \ac{ml} methods, and in particular, \acp{dnn}, have been the focus of extensive research in recent years due to their empirical success in various applications, including image processing and speech processing \cite{LeCun:15}. 
	There are several benefits in using \ac{ml} schemes over traditional model-based approaches: 
	First, \ac{ml} methods are independent of the underlying stochastic model, and thus can operate efficiently in scenarios where this model is unknown or its parameters cannot be accurately estimated.
	Second, when the underlying model is extremely complex, \ac{ml} algorithms have demonstrated the ability to extract and disentangle meaningful semantic information from the observed data \cite{Bengio:09}, a task which is very difficult to carry out using traditional model-based approaches. 
	 Finally, \ac{ml} techniques often lead to faster convergence compared to iterative model-based approaches, even when the model is known \cite{Gregor:10, Khobahi:19}.

	Recent years have witnessed growing interest in the application of \acp{dnn} for receiver design. Due to the large amount of recent works on this topic, we only discuss a few representative papers; detailed surveys can be found in \cite{Oshea:17, Simeone:18, Mao:18, Gunduz:19}: The work \cite{Nachmani:17} used deep learning for decoding linear codes. Decoding of structured codes using   \acp{dnn} was considered in \cite{Gruber:17}, while symbol recovery in \ac{mimo} systems was treated in \cite{Wiesel:17a}. The work \cite{Caciularu:18} used variational autoencoders for equalizing linear multipath channels. Sequence detection using bi-directional  \acp{rnn} was proposed in \cite{Farsad:18}, while \cite{Liang:18} considered a new \ac{ml}-based channel decoder by combining convolutional neural networks with belief propagation. \ac{rnn} structures for decoding sequential codes were studied in \cite{Kim:18}.  
	The work \cite{Wang:96} proposed a neural network architecture that enables parallel implementation of the Viterbi algorithm with binary signals in hardware.
	\ac{dnn}-based \ac{mimo} receivers for mitigating the effect of low-resolution quantization were studied in \cite{Shlezinger:19}. 
	These  approaches were shown to yield good performance when sufficient training is available. 
	However, previous applications of \acp{dnn} typically treat the network as a black box, hoping to achieve the desired performance by using sufficient training and relying on methods that were developed to treat other tasks such as image processing. This gives rise to the question of whether additional gains, either in performance, complexity, or training size, can be achieved by combining channel-model-based methods, such as the Viterbi algorithm, with \ac{ml}-based techniques.

	In this work, we design and study \ac{ml}-based symbol detection for finite-memory causal  channels based on the Viterbi algorithm. Our design is inspired by deep unfolding, which is a common method for obtaining \ac{ml} architectures from model-based iterative algorithms \cite{Gregor:10,Hershey:14}. Unfolding was shown to yield efficient and reliable \acp{dnn} for applications such as sparse recovery \cite{Gregor:10}, recovery from one-bit measurements \cite{Khobahi:19},  matrix factorization \cite{Sprechmann:14}, image deblurring \cite{Li:19}, and robust principal component analysis \cite{Solomon:18}.
	However, there is a fundamental difference between our approach and conventional unfolding: 
	 The main rationale of unfolding is to 	convert each iteration of the algorithm into a layer, namely, to design a \ac{dnn} in light of a model-based algorithm, or alternatively, to {\em integrate the algorithm into the \ac{dnn}}. Our approach to symbol detection implements the Viterbi channel-model-based algorithm, while only removing its channel-model-dependence by replacing the \ac{csi}-based computations with dedicated \acp{dnn}, i.e., we {\em integrate \ac{ml} into the Viterbi algorithm}. 
	
	In particular, we propose ViterbiNet, which is an \ac{ml}-based symbol detector integrating deep learning into the Viterbi algorithm. The resulting system approximates the mapping carried out by the channel-model-based Viterbi method in a data-driven fashion without requiring \ac{csi}, namely, without  knowing the exact channel input-output statistical relationship. ViterbiNet combines established properties of \ac{dnn}-based sequence classifiers to implement the specific computations in the Viterbi algorithm that are \ac{csi}-dependent. 
	Unlike direct application of \ac{ml} algorithms, the resulting detector is  capable of exploiting the underlying Markovian structure of finite-memory causal channels in the same manner as the Viterbi algorithm. 
	ViterbiNet	consists of a simple network architecture which can be trained with a relatively small number of training samples. 
	
	Next, we propose a method for adapting ViterbiNet to dynamic channel conditions online without requiring new training data every time the statistical model of the channel changes. Our approach is based on meta-learning \cite{Lemke:15}, namely, automatic learning and adaptation of \acp{dnn}, typically by acquiring training from related tasks. The concept of meta-learning was recently used in the context of \ac{ml} for communications in the work \cite{Park:19}, which used pilots from neighboring devices as meta-training in an \ac{iot} network.   Our proposed approach utilizes channel coding to generate meta-training from each decoded block. In particular, we use the fact that \ac{fec} codes can compensate for decision errors as a method for generating accurate training from the decoded block. This approach exploits both the inherent structure of coded communication signals, as well as the ability of ViterbiNet to train with the relatively small number of samples obtained from decoding a single block. 
	
	Our numerical evaluations demonstrate that, when the training data obeys the same statistical model as the test data, ViterbiNet achieves roughly the same performance as the \ac{csi}-based Viterbi algorithm. Furthermore, when ViterbiNet is trained for a variety of different channels, it notably outperforms the Viterbi algorithm operating with the same level of \ac{csi} uncertainty.	
	It is also illustrated that ViterbiNet performs well in complex channel models, where the Viterbi algorithm is extremely difficult to implement even when \ac{csi} is available.
	In the presence of block-fading channels, we show that by using our proposed online training scheme, ViterbiNet is capable of tracking the varying channel conditions and approaches the performance achievable with the Viterbi detector, which requires accurate instantaneous \ac{csi} for each block.  
	Our results demonstrate that reliable, efficient, and robust \ac{ml}-based communication systems can be realized by integrating \ac{ml} into existing \ac{csi}-based techniques and accounting for the inherent structure of digital communication signals.

 	
 	The rest of this paper is organized as follows: In Section~\ref{sec:Preliminaries} we present the system model and review the Viterbi algorithm. Section~\ref{sec:ViterbiNet}  proposes ViterbiNet: A receiver architecture that implements Viterbi detection using \acp{dnn}. Section~\ref{sec:BlockFading} presents a method for training ViterbiNet online, allowing it to track block-fading channels. Section~\ref{sec:Sims} details numerical training and performance results of ViterbiNet, and Section~\ref{sec:Conclusions} provides concluding remarks.

 	\smallskip
 	Throughout the paper, we use upper case letters for \acp{rv}, e.g. $X$.
 	Boldface lower-case letters denote vectors, e.g., ${\myVec{x}}$ is a deterministic vector, and $\myVec{X}$ is a random vector, and
 	the $i$th element of ${\myVec{x}}$ is written as $\left( {\myVec{x}}\right) _i$. 
 	The \ac{pdf} of an \ac{rv} $X$ evaluated at $x$ is denoted $\Pdf{X}(x)$,  
 	%
 	%
 	$\mySet{Z}$ is the set of integers,  	$\mySet{N}$ is the set of positive integers, 
 	 $\mySet{R}$ is the set of real numbers, and $(\cdot)^T$ is the transpose operator. 
 	All logarithms are taken to basis 2. 
 	Finally, for any sequence, possibly multivariate, $\myVec{y}[i]$, $i \in \mySet{Z}$, and integers $b_1 < b_2$,  $\myVec{y}_{b_1}^{b_2}$ is the column vector $\left[\myVec{y} ^T[b_1],\ldots, \myVec{y}^T[b_2] \right]^T$ and $\myVec{y}^{b_2} \equiv \myVec{y}_{1}^{b_2}$.
	\vspace{-0.2cm}
	\section{System Model and Preliminaries}
	\label{sec:Preliminaries}
	\vspace{-0.1cm}
	
	\subsection{System Model}
	\label{subsec:Model}
\begin{figure}
	\centering
	\includefig{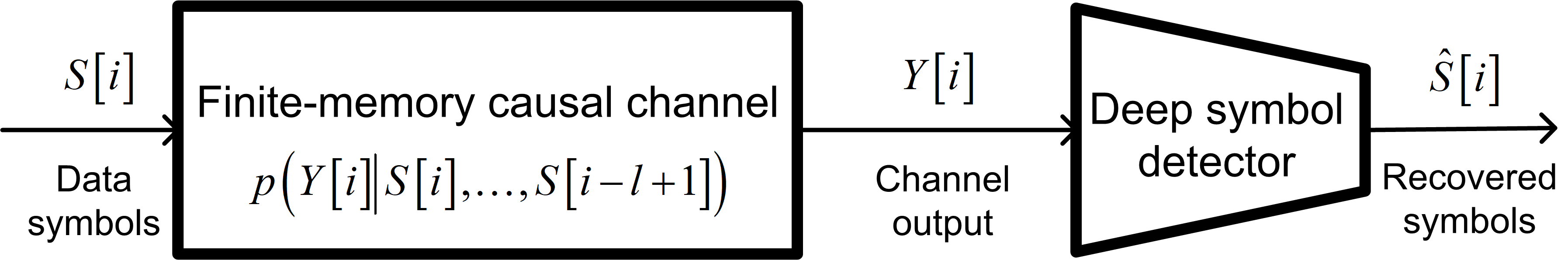} 
	\caption{System model.}
	\label{fig:BasicModel2}
\end{figure}

	\vspace{-0.1cm}
	We consider the problem of recovering a block of $\Blklen$ symbols transmitted over a finite-memory stationary causal channel. Let $S[i] \in \mySet{S}$ be the symbol transmitted at time index $i \in \{1,2,\ldots, \Blklen\}\triangleq \Blkset$, where each symbol is uniformly distributed over a set of $\CnstSize$ constellation points, thus $| \mySet{S}| = \CnstSize$. We use $Y[i]\in \mySet{Y}$ to denote the channel output at time index $i$. Since the channel is causal and has a finite memory, $Y[i] $ is given by some stochastic mapping of $\myVec{S}_{i-\Mem+1}^{i}$, where $\Mem$ denotes the memory of the channel, assumed to be smaller than the blocklength, i.e.,  $\Mem < \Blklen$.
	Consequently, the conditional \ac{pdf} of the channel output given the channel input satisfies   
	\begin{equation}
	\label{eqn:ChModel1}
	\Pdf{\myVec{Y}_{k_1}^{k_2} | \myVec{S}^{ \Blklen}}\left(\myVec{y}_{k_1}^{k_2} | \myVec{s}^{ \Blklen} \right)  = 
	\prod\limits_{i\!=\!k_1}^{k_2}\Pdf{Y[i]  | \myVec{S}_{i\!-\!\Mem\!+\!1}^{i}}\left( y[i]  | \myVec{s}_{i\!-\!\Mem\!+\!1}^{i}\right), 
	\end{equation}
	for all  $k_1, k_2\in \Blkset$ such that $ k_1 \le k_2$. The fact that the channel is stationary implies that for each $y \in \mySet{Y}$, $\myVec{s} \in \mySet{S}^\Mem$, the conditional \ac{pdf} $\Pdf{Y[i]  | \myVec{S}_{i-\Mem+1}^{i}}\left( y  | \myVec{s}\right)$ does not depend on the  index $i$. An illustration of the system is depicted in Fig. \ref{fig:BasicModel2}.

	Our goal is to design a \ac{dnn} architecture for recovering  $\myVec{S}^{ \Blklen}$ from the channel output $\myVec{Y}^{ \Blklen}$.  In particular, in our model the receiver assumes that the channel is stationary, causal, and has finite memory $\Mem$, namely, that the input-output statistical relationship is of the form \eqref{eqn:ChModel1}. The receiver also knows the constellation $\mySet{S}$. We do not assume that the receiver knows the conditional \ac{pdf} $\Pdf{Y[i]  | \myVec{S}_{i\!-\!\Mem\!+\!1}^{i}}\left( y  | \myVec{s}\right)$, i.e., the receiver does not have this \ac{csi}.  
	The optimal  detector for finite-memory channels, based on which we design our network, is the  Viterbi algorithm \cite{Viterbi:67}. 
	Therefore, as a preliminary step to designing the \ac{dnn}, we  review Viterbi detection  in the following subsection.

	\vspace{-0.2cm}
	\subsection{The Viterbi Detection Algorithm}
	\label{subsec:Viterbi}
	\vspace{-0.1cm}
The following description of the Viterbi algorithm is based on \cite[Ch. 3.4]{Tse:05}. 
Since the constellation points are equiprobable, the optimal decision rule in the sense of minimal probability of error is the maximum likelihood decision rule, namely, for a given channel output $\myVec{y}^{ \Blklen}$, the estimated output is given by  
\begin{align}
\hat{\myVec{s}}^{ \Blklen}\left( \myVec{y}^{ \Blklen}\right)  
&\triangleq \mathop{\arg \max}_{\myVec{s}^{ \Blklen} \in \mySet{S}^\Blklen } \Pdf{\myVec{Y}^{ \Blklen} | \myVec{S}^{ \Blklen}}\left( {\myVec{y}^{ \Blklen} | \myVec{s}^{ \Blklen}}\right)  \notag \\
&= \mathop{\arg \min}_{\myVec{s}^{ \Blklen} \in \mySet{S}^\Blklen } -\log \Pdf{\myVec{Y}^{ \Blklen} | \myVec{S}^{ \Blklen}}\left( \myVec{y}^{ \Blklen} | \myVec{s}^{ \Blklen}\right) .
\label{eqn:ML1}
\end{align}
By defining  
\begin{equation}
c_i\left( \myVec{s}\right)  \triangleq- \log \Pdf{\myVec{Y}[i]  | \myVec{S}_{i-\Mem+1}^{i}}\left( \myVec{y}[i]  | \myVec{s}\right), \qquad \myVec{s} \in \mySet{S}^{\Mem}, 
\label{eqn:CondProb1}
\end{equation} 
it follows from \eqref{eqn:ChModel1} that the log-likelihood function in \eqref{eqn:ML1} can be written as
\begin{equation}
\log \Pdf{\myVec{Y}^{ \Blklen} | \myVec{S}^{ \Blklen}} \left( \myVec{y}^{ \Blklen} | \myVec{s}^{ \Blklen}\right) 
= \sum\limits_{i=1}^{\Blklen } c_i\left( \myVec{s}_{i-\Mem+1}^{i}\right),
\label{eqn:ML2}
\end{equation}
and the optimization problem \eqref{eqn:ML1} becomes
\begin{align}
\hat{\myVec{s}}^{ \Blklen}\left( \myVec{y}^{ \Blklen}\right)   
&= \mathop{\arg \min}_{\myVec{s}^{ \Blklen} \in \mySet{S}^\Blklen }\sum\limits_{i=1}^{\Blklen } c_i\left( \myVec{s}_{i-\Mem+1}^{i}\right).
\label{eqn:ML3}
\end{align}

The optimization problem \eqref{eqn:ML3} can be solved recursively using dynamic programming, by treating the possible combinations of transmitted symbols at each time instance as {\em states} and iteratively updating a {\em path cost} value for each state, The resulting recursive solution, which is known as the Viterbi algorithm, is given below as Algortihm~\ref{alg:Algo1}. 
\begin{algorithm}
	\caption{ The Viterbi Algorithm}
	\label{alg:Algo1}
	\begin{algorithmic}[1]
		\STATE \underline{Input}: Block of channel outputs $\myVec{y}^{ \Blklen}$, where $\Blklen > \Mem$. 
		\STATE \underline{Initialization}: Set $k=1$, and fix the initial path cost $\tilde{c}_{0}\left( \tilde{\myVec{s}}\right) =  0$, for each state $\tilde{\myVec{s}} \in \mySet{S}^\Mem$.
		\STATE \label{stp:MF1} For each state $\tilde{\myVec{s}} \in \mySet{S}^\Mem$, compute the path cost via 
	$	\tilde{c}_k\left( \tilde{\myVec{s}}\right) = \mathop{\min}\limits_{\myVec{u} \in  \mySet{S}^\Mem: \myVec{u}_2^{\Mem} =  \tilde{\myVec{s}}^{\Mem - 1}} \left(\tilde{c}_{k-1}\left( \myVec{u}\right) + {c}_k\left( \tilde{\myVec{s}}\right) \right)$.  
		\STATE \revision{If $k \ge \Mem $, set $\left( \hat{\myVec{s}}\right)_{k\! -\! \Mem\! +\! 1} \!:=\! \left( \tilde{\myVec{s}}_k^{\rm o}\right)_1$, where 
			$\tilde{\myVec{s}}^{\rm o}_k \!=\! \mathop{\arg \min}\limits_{\tilde{\myVec{s}} \in \mySet{S}^\Mem}\tilde{c}_k\left( \tilde{\myVec{s}}\right)$. }  
%
		\STATE Set $k := k+1$. If $k \le \Blklen $ go to Step \ref{stp:MF1}.	
		
		\STATE  \underline{Output}: decoded output $\hat{\myVec{s}}^{ \Blklen}$,  where	\revision{$\hat{\myVec{s}}_{\Blklen-\Mem + 1}^{ \Blklen} := \tilde{\myVec{s}}_{{ \Blklen}}^{\rm o}$.}
	\end{algorithmic}
\end{algorithm} 

In addition to its ability to achieve the minimal error probability \cite{Forney:73}, \textcolor{NewColor}{the} Viterbi algorithm has several major advantages:
\begin{enumerate}[label={\em A\arabic*}]
	\item \label{itm:Adv1} The algorithm solves \eqref{eqn:ML1} at a computational complexity that is linear in the blocklength $\Blklen$. For comparison, the computational complexity of solving \eqref{eqn:ML1} directly grows exponentially with $\Blklen$.  	
	\item \label{itm:Adv2} The algorithm produces estimates sequentially on run-time. In particular, while in \eqref{eqn:ML1}  the estimated output $\hat{\myVec{s}}^{ \Blklen}$ is computed using the entire received block $\myVec{y}^{ \Blklen}$, Algorithm \ref{alg:Algo1} computes $\hat{s}[i]$  once $y[i + \Mem - 1]$ is received.  
\end{enumerate}

In order to implement Algorithm \ref{alg:Algo1}, one must be able to compute  ${c}_i\left( {\myVec{s}}\right)$ of \eqref{eqn:CondProb1} for all $i \in \Blkset$ and for each $\myVec{s} \in \mySet{S}^\Mem$. Consequently, the conditional \ac{pdf} of the channel, which we refer to as full \ac{csi}, must be explicitly known. As discussed in the introduction, obtaining full \ac{csi} may be extremely difficult in rapidly changing channels and may also require a large training overhead. 
In the following section we propose ViterbiNet,  an \ac{ml}-based symbol decoder based on the Viterbi algorithm, that does not require \ac{csi}. 
	
\vspace{-0.2cm}
\section{ViterbiNet}
\label{sec:ViterbiNet}
\vspace{-0.1cm}
\subsection{Integrating \ac{ml} into the Viterbi Algorithm}
\label{subsec:Derivation}
\vspace{-0.1cm}
%
%
%
In order to integrate \ac{ml} into the Viterbi Algorithm, we note that  \ac{csi} is  required in Algorithm~\ref{alg:Algo1} only in Step~\ref{stp:MF1} to compute the log-likelihood function $ c_i (\myVec{s})$. 
Once  $ c_i (\myVec{s})$ is computed for each $\myVec{s} \in \mySet{S}^\Mem$, the Viterbi algorithm only requires knowledge of the memory length $\Mem$ of the channel. This requirement is much easier to satisfy compared to knowledge of the exact channel input-output statistical relationship, i.e., full \ac{csi}. 
In fact, a common practice when using the model-based Viterbi algorithm is to limit the channel memory by filtering the channel output prior to symbol detection \cite{Falconer:73} in order to reduce computational complexity. The approach in \cite{Falconer:73}, which assumes that the channel is modeled as a \ac{lti} filter, is an example of how the channel memory used by the algorithm can be fixed in advance, regardless of the true memory length of the underlying channels. 
Furthermore, estimating the channel memory in a data-driven fashion without a-priori knowledge of the underlying model is a much simpler task compared to symbol detection, and can be reliably implemented using standard correlation peak based estimators, see, e.g., \cite{Nguyen:07}. 
We henceforth assume that the receiver has either accurate knowledge or a reliable upper bound on $\Mem$, and leave the extension of the decoder to account for unknown memory length to future exploration.  

\begin{figure}
	\centering
	\includefig{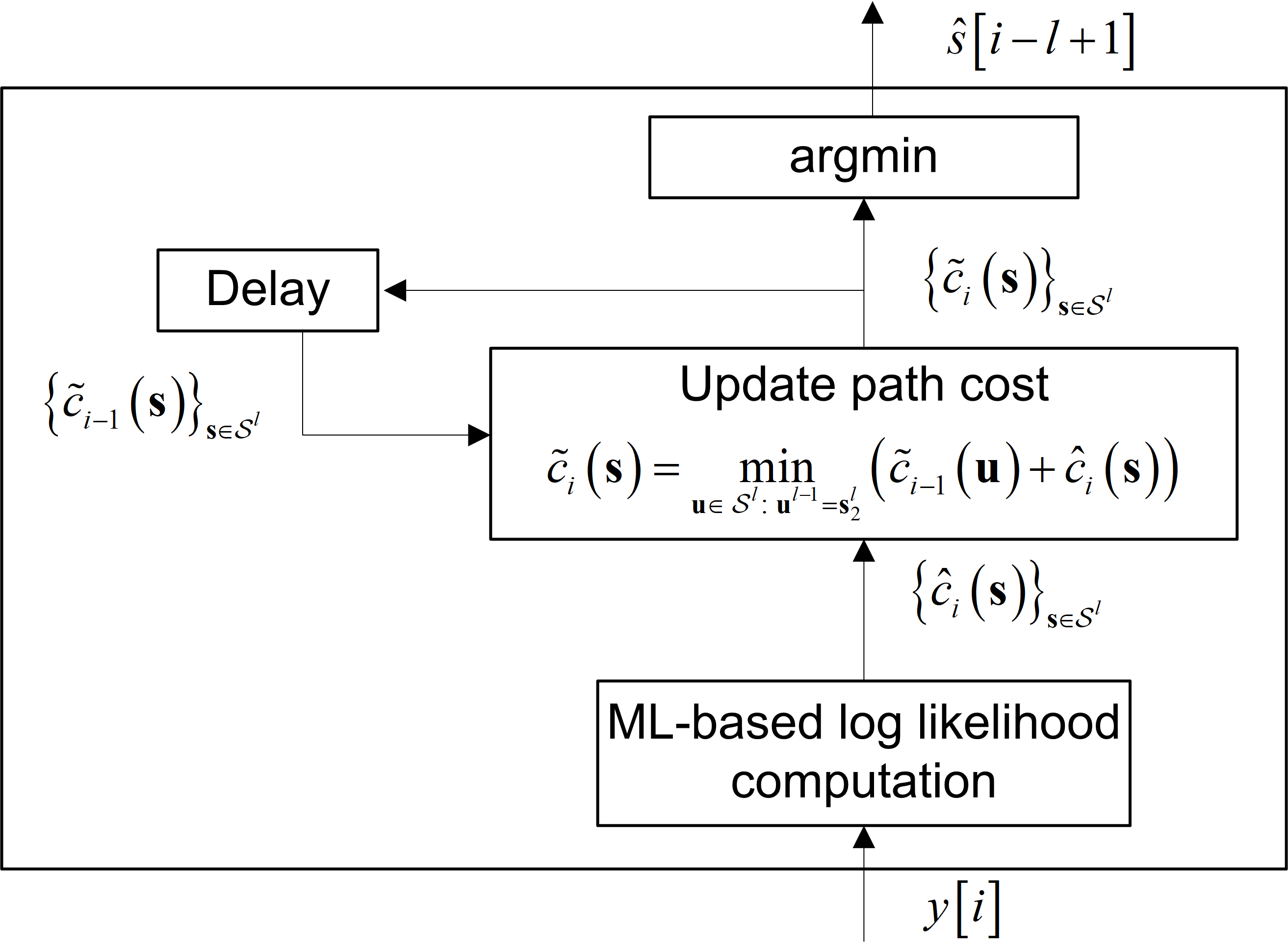} 
	\caption{Proposed \ac{dnn}-based Viterbi decoder.}
	\label{fig:DNNSystem}
\end{figure}

Since the channel is stationary, it holds by \eqref{eqn:CondProb1} that if $y[i] = y[k]$ then $ c_i (\myVec{s}) = c_k(\myVec{s})$, for each $\myVec{s} \in \mySet{S}^\Mem$, and  $i,k \in \Blkset$. Consequently, the log-likelihood function  $ c_i (\myVec{s})$ depends only on the values of $y[i]$ and of $\myVec{s}$, and not on the time index $i$. 
Therefore, in our proposed structure we replace the explicit computation of the log-likelihood \eqref{eqn:CondProb1} with an \ac{ml}-based system that learns to evaluate the cost function from the training data. In this case, the input of the system is $y[i]$ and the output is an estimate of $c_i({\myVec{s}})$, denoted $\hat{c}_i({\myVec{s}})$, for each $\myVec{s}\in\mySet{S}^{\Mem}$. The rest of the Viterbi algorithm remains intact, namely, the detector implements Algorithm~\ref{alg:Algo1} while using \ac{ml} techniques to compute the log-likelihood function $c_i({\myVec{s}})$.   The proposed architecture is illustrated in Fig. \ref{fig:DNNSystem}.

A major challenge in implementing a network capable of computing $c_i(\myVec{s})$ from $y[i]$ stems from the fact that, by \eqref{eqn:CondProb1}, $c_i(\myVec{s})$ represents the log-likelihood of $Y[i] = y[i]$ given $\myVec{S}_{i-\Mem+1}^i = \myVec{s}$. However,  \acp{dnn} trained to minimize the cross entropy loss typically output the conditional distribution of   $\myVec{S}_{i-\Mem+1}^i = \myVec{s}$ given  $Y[i] = y[i]$, i.e., $\Pdf{\myVec{S}_{i-\Mem+1}^i| Y[i]}\left( \myVec{s}| y[i]\right)$. Specifically, classification \acp{dnn} with input $Y[i] = y[i]$, which typically takes values in a discrete set, or alternatively, is discretized by binning a continuous quantity as in  \cite{Oord:16a,Oord:16}, 
output the  distribution of the label $\myVec{S}_{i-\Mem+1}^i$ conditioned on that input $Y[i] = y[i]$, and not the distribution of the input conditioned on all possible values of the label  $\myVec{S}_{i-\Mem+1}^i$.
For this reason, the previous work \cite{Farsad:18} used a \ac{dnn} to approximate the \ac{map} detector by considering  $\Pdf{S[i]| \myVec{Y}_{i-\Mem+1}^i}\left( s| \myVec{y}_{i-\Mem+1}^i\right)$.
It is emphasized that the quantity which is needed for the Viterbi algorithm is the conditional \ac{pdf} $\Pdf{Y[i]|\myVec{S}_{i-\Mem+1}^i}\left(  y[i]| \myVec{s}\right)$, and not the conditional distribution $\Pdf{\myVec{S}_{i-\Mem+1}^i| Y[i]}\left( \myVec{s}| y[i]\right)$.  
Note, for example, that while $\sum_{\myVec{s}\in\mySet{S}^\Mem} \Pdf{\myVec{S}_{i-\Mem+1}^i| Y[i]}\left( \myVec{s}| y[i]\right) = 1$, for the desired conditional \ac{pdf}  $\sum_{\myVec{s}\in\mySet{S}^\Mem} \Pdf{Y[i]|\myVec{S}_{i-\Mem+1}^i}\left(  y[i]| \myVec{s}\right) \neq 1$ in general.  
Therefore, outputs generated using conventional \acp{dnn} with a softmax output layer are not applicable. 
The fact that  Algorithm \ref{alg:Algo1} specifically uses the conditional \ac{pdf}   $\Pdf{Y[i]|\myVec{S}_{i-\Mem+1}^i}\left(  y[i]| \myVec{s}\right)$ instead of $\Pdf{\myVec{S}_{i-\Mem+1}^i| Y[i]}\left( \myVec{s}| y[i]\right)$ allows it to exploit the Markovian nature of the channel, induced by the finite memory in \eqref{eqn:ChModel1}, resulting in the advantages \ref{itm:Adv1}-\ref{itm:Adv2} discussed in Subsection \ref{subsec:Viterbi}.

%

In order to tackle this difficulty, we recall that by Bayes' theorem, as the channel inputs are equiprobable, the desired conditional \ac{pdf} $\Pdf{Y[i]|\myVec{S}_{i-\Mem+1}^i}\left(  y| \myVec{s}\right)$ can be written as 
\begin{equation}
\label{eqn:Bayes}
\Pdf{Y[i]|\myVec{S}_{i-\Mem+1}^i}\left(  y| \myVec{s}\right) = \frac{\Pdf{\myVec{S}_{i-\Mem+1}^i| Y[i]}\left( \myVec{s}| y\right)\cdot\Pdf{Y[i]}(y)}{ \CnstSize^{-\Mem}}.
\end{equation} 
Therefore, given estimates of  $\Pdf{Y[i]}(y[i])$ and of  $\Pdf{\myVec{S}_{i-\Mem+1}^i| Y[i]}\left( \myVec{s}| y[i]\right)$ for each $\myVec{s} \in \mySet{S}^\Mem$, the log-likelihood function $c_i(\myVec{s})$ can be recovered using \eqref{eqn:Bayes} and \eqref{eqn:CondProb1}.

\begin{figure}
		\centering
		\includefig{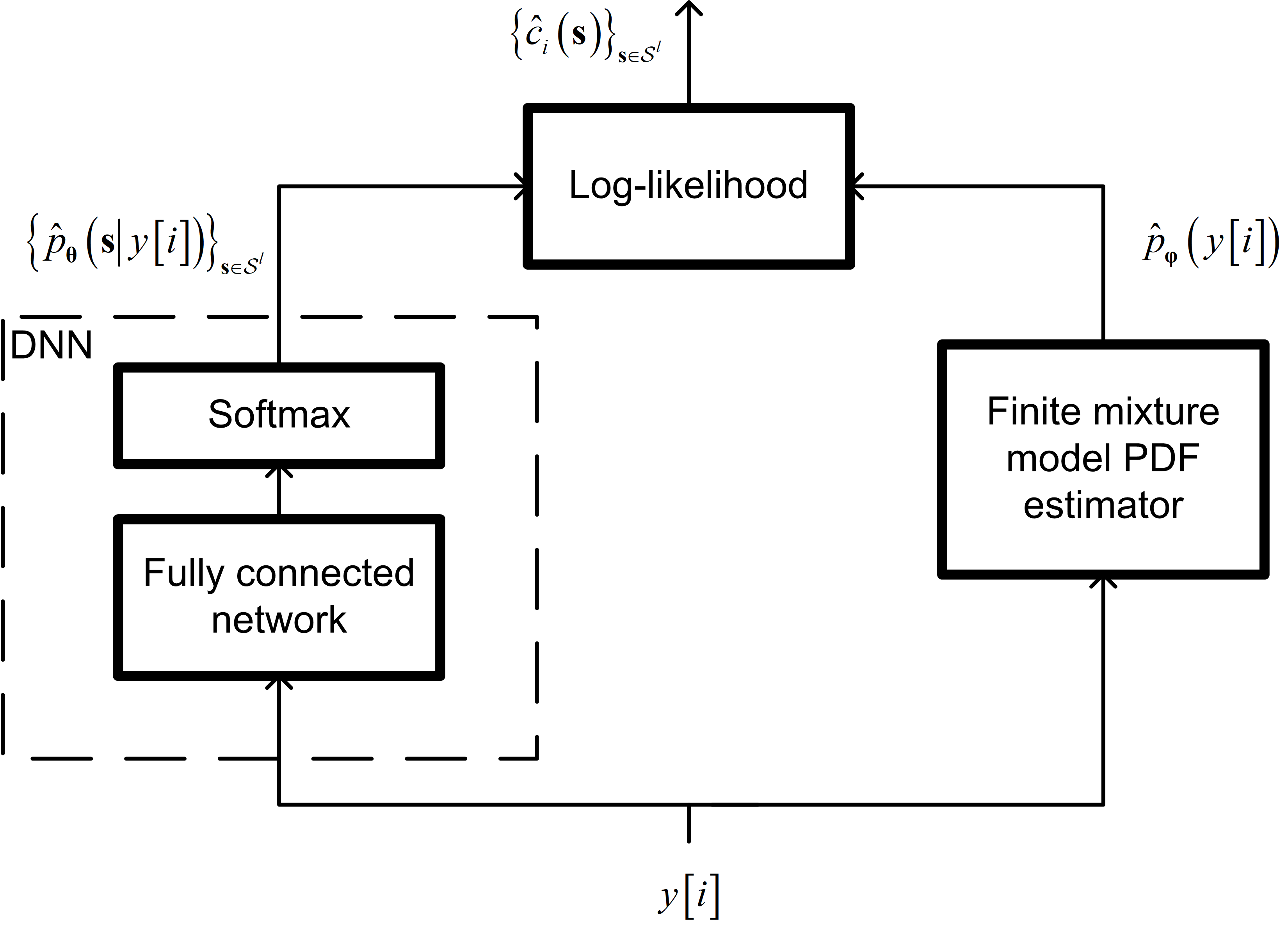}
		\caption{\ac{ml}-based log-likelihood computation.}
		\label{fig:NetworkArchitecture}
	\end{figure}

A parametric estimate of $\Pdf{\myVec{S}_{i-\Mem+1}^i| Y[i]}\left( \myVec{s}| y[i]\right)$, denoted  $\PdfEst{\myVec{\theta}}\left( \myVec{s}| y[i]\right)$, can be reliably obtained from training data using standard classification \acp{dnn}  with a softmax output layer. The marginal \ac{pdf} of $Y[i]$ may be estimated from the training data using conventional kernel density estimation methods \cite{Rosenblatt:56}. Furthermore, the fact that $Y[i]$ is a stochastic mapping of $\myVec{S}_{i-\Mem+1}^i$ implies that its distribution can be approximated as a mixture model of $\CnstSize^\Mem$ kernel functions \cite{McLachlan:04}. Consequently, a parametric estimate of $\Pdf{Y[i]}(y[i])$, denoted $\PdfEst{\myVec{\varphi}}\left(  y[i]\right)$, can be obtained from the training data using mixture density networks \cite{Bishop:94}, \ac{em}-based algorithms \cite[Ch. 2]{McLachlan:04}, or any other finite mixture model fitting method. The resulting \ac{ml}-based log-likelihood computation is illustrated in Fig. \ref{fig:NetworkArchitecture}.

\label{txt:Complex}	
\revision{The \ac{ml}-based log-likelihood computation depicted in Fig. \ref{fig:NetworkArchitecture} is applicable for both real-valued channels, where $\mySet{Y}$ and $\mySet{S}$ are subsets of $\mySet{R}$, as well as complex channels, in which the elements of $\mySet{Y}$ and $\mySet{S}$ take complex values. In particular, for complex-valued channels, the input to the classification \ac{dnn} should be a real vector consisting of the real and imaginary parts of $y[i]$, as \acp{dnn} typically operate on real-valued quantities. Furthermore, the finite mixture model \ac{pdf} estimator should use kernel functions for complex distributions, e.g., complex Gaussian mixture. Except for these modifications, the architecture of ViterbiNet is invariant as to whether the channel is real or complex-valued.}	
	
\vspace{-0.2cm}
\subsection{Discussion}
\label{subsec:Discussion}
\vspace{-0.1cm}
The architecture of ViterbiNet is based on the Viterbi algorithm, which achieves the minimal probability of error for communications over finite-memory causal channels. Since ViterbiNet is data-driven, it 
learns the log-likelihood function from training data without requiring full \ac{csi}. When properly trained, the proposed \ac{dnn} is therefore expected to approach the performance achievable with the conventional {\ac{csi}}-based Viterbi algorithm, as numerically demonstrated in Section \ref{sec:Sims}.  \revision{Unlike direct applications of classification \acp{dnn}, which produce a parametric estimate of the conditional distribution of the transmitted symbols given the channel output, ViterbiNet is designed to estimate the conditional distribution of the channel output given the transmitted symbols. This conditional distribution, which encapsulates the Markovian structure of finite-memory channels, is then used to obtain the desired log-likelihoods, as discussed in the previous subsection.  The estimation of these log-likelihoods, which are not obtained by simply applying a classification \ac{dnn} to the channel output, is then exploited to recover the transmitted symbols using the same operations carried out by the conventional Viterbi algorithm.}

ViterbiNet  replaces the \ac{csi}-based parts of the Viterbi algorithm with an \ac{ml}-based scheme.  Since this aspect of the channel model is relatively straightforward to learn, ViterbiNet can utilize a  simple and standard \ac{dnn} architecture.   
The simple \ac{dnn} structure implies that the network can be trained quickly with a relatively small number of training samples, as also observed in the numerical study in Section \ref{sec:Sims}. This indicates the potential of the proposed architecture to adapt on runtime to channel variations with minimal overhead, possibly using pilot sequences periodically embedded in the transmitted frame for online training. 

\label{txt:BCJR}
\revision{
The approach used in designing ViterbiNet, namely, maintaining the detection scheme while replacing its channel-model-based computations with dedicated \ac{ml} methods, can be utilized for implementing additional algorithms in communications and signal processing in a data-driven manner. For instance, in \cite{Shlezinger:19b}, we showed how this approach can allow a \ac{mimo} receiver to learn to implement the soft iterative interference cancellation scheme of \cite{Choi:00} from a small training set. In particular, the \ac{ml}-based computation of the conditional probability used in ViterbiNet to carry out Viterbi detection can be utilized to realize other trellis or factor graph based detection schemes, such as the BCJR algorithm \cite{Bahl:74}, by properly modifying the processing of these estimates. We leave the study of these alternative data-driven schemes for future investigation.}

The fact that ViterbiNet is based on the Viterbi algorithm implies that it also suffers from some of its drawbacks. For example, when the constellation size  $\CnstSize$ and the channel memory $\Mem$ grow large, the Viterbi algorithm becomes computationally complex due to the need to compute the log-likelihood for each of the possible $\CnstSize^\Mem$ different values of $\myVec{s}$. Consequently, the complexity of ViterbiNet is expected to grow exponentially as $\CnstSize$ and $\Mem$ grow, since the label space of the \ac{dnn} grows exponentially. It is noted that greedy schemes for reducing the complexity of the Viterbi algorithm, such as beam search \cite{Lingyun:04} and reduced-state equalization \cite{Gerstacker:02}, were shown to result in minimal performance degradation, and we thus expect that these methods can inspire similar modifications to ViterbiNet, facilitating its application with large $\CnstSize$ and $\Mem$. We leave the research into reducing the complexity of ViterbiNet through such methods to future investigation.

\vspace{-0.2cm}
\section{Extension to Block-Fading Channels}
\label{sec:BlockFading}
\vspace{-0.1cm}
A major challenge associated with using \ac{ml} algorithms in communications stems from the fact that, in order to exploit the ability of \acp{dnn} to learn an underlying model, the network should operate under the same (or a closely resembling) statistical relationship for which it was trained. Many communication channels, and particularly wireless channels, are dynamic by nature, and are commonly modeled as {\em block-fading} \cite{Hassibi:03}. In such channels, each transmitted block may undergo a different statistical transformation. 
In Section \ref{sec:Sims}, we show that ViterbiNet is capable of achieving relatively good performance when trained using samples taken from a variety of channels rather than from the specific channel for which it is tested. Consequently, a possible approach to use ViterbiNet in block-fading channels is to train the system using samples acquired from a large variety of channels. However, such an approach requires a large amount of training data in order to cover a multitude of different channel conditions, and is also expected to result in degraded performance, especially when tested under a channel model that deviates significantly from the channels used during training. 
This motivates us to extend ViterbiNet by allowing the receiver to track and adjust to time-varying channel conditions in real-time, by exploiting the inherent structure of digital communication signals and, particularly, of coded communications. Broadly speaking, we utilize the fact that in coded systems, the receiver is capable of identifying and correcting symbol detection errors. The identification and correction of these errors can be used to retrain ViterbiNet online using meta-learning principles \cite{Lemke:15}, allowing it to track channel variations. To present this approach, we first extend the channel model of Subsection \ref{subsec:Coded} to account for coded communications and block-fading channels. Then, we detail in Subsection \ref{subsec:Online} how ViterbiNet can exploit the use of channel coding for online training. 

\subsection{Coded Communications over Block-Fading Channels}
\label{subsec:Coded}
\vspace{-0.1cm} 
In coded communications, each block of $\Blklen$ transmitted channel symbols represents a channel codeword conveying $\Codlen$ bits. During the $j$th block, $j \in \mySet{N}$, a vector of bits, denoted $\myVec{B}^\Codlen\Blkidx \in \{0,1\}^\Codlen$, is encoded and modulated into the symbol vector $\myVec{S}^\Blklen\Blkidx$ which is transmitted over the channel. The purpose of such channel coding is to facilitate the recovery of the information bits at the receiver. 
A channel code can consist of both \ac{fec} codes, such as \ac{rs} codes, as well as error detection codes, such as checksums and cyclic redundancy checks. The receiver uses the recovered symbols $\hat{\myVec{S}}^\Blklen\Blkidx$ to decode the information bits  $\hat{\myVec{B}}^\Codlen\Blkidx \in \{0,1\}^\Codlen$. The presence of \ac{fec} codes implies that even when some of the symbols are not correctly recovered, i.e., $\myVec{S}^\Blklen\Blkidx \neq \hat{\myVec{S}}^\Blklen\Blkidx$, the information bits can still be perfectly decoded. Error detection codes allow the receiver to detect whether the information bits were correctly recovered and estimate the number of bit errors, i.e., the Hamming distance between $\myVec{B}^\Codlen\Blkidx$ and $\hat{\myVec{B}}^\Codlen\Blkidx$, denoted $\epsilon\Blkidx$. 

\begin{figure}
	\centering
	\includefig{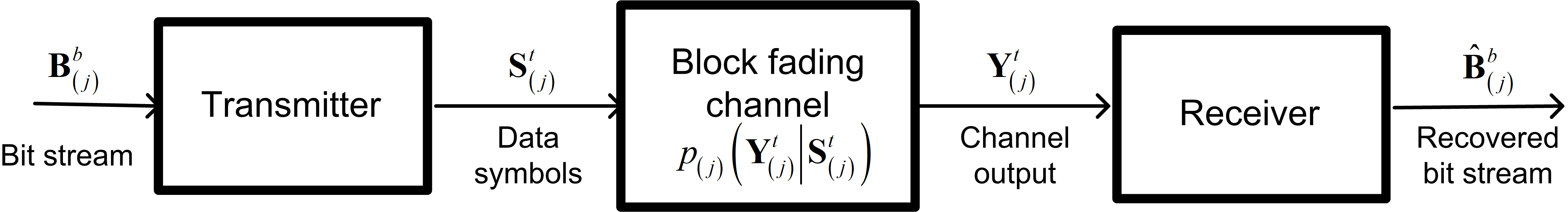}
	\caption{Coded communications over block-fading channels.}
	\label{fig:BlockFadingModel2}
\end{figure}

In block-fading channels, each block of transmitted channel symbols  undergoes a different channel. In particular, for the $j$th block, $j \in \mySet{N}$, the conditional distribution of the channel output $\myVec{Y}^\Blklen\Blkidx$ given the input  $\myVec{S}^\Blklen\Blkidx$ represents a finite-memory causal stationary channel as described in Subsection \ref{subsec:Model}. However, this conditional distribution can change between different blocks, namely, it depends on the block index $j$. An illustration of this channel model is depicted in Fig. \ref{fig:BlockFadingModel2}. Block-fading channels model dynamic environments in which the transmitted bursts are shorter than the coherence duration of the channel, faithfully representing many scenarios of interest in wireless communications \cite{Hassibi:03}.  

Since each block undergoes a different channel input-output relationship, the receiver must be able to track and adapt to the varying channel conditions in order to optimize its performance. In the next subsection, we show how the fact that ViterbiNet requires relatively small training sets can be combined with the presence of coded communications such that the receiver can track channel variations online.

\vspace{-0.2cm}
\subsection{ViterbiNet with Online Training}
\label{subsec:Online}
\vspace{-0.1cm}
We next discuss how coded communications can allow ViterbiNet to train and track channel variations online. Our proposed approach is inspired by concepts from meta-learning \cite{Lemke:15}, which is a field of research focusing on self-teaching \ac{ml} algorithms, as well as from decision-directed adaptive filter theory, which considers the blind adaptation of adjustable filters based on some decision mechanism, see, e.g., \cite{Shlezinger:14}. 

	\begin{figure*}
	\centering
	\includegraphics[width = \linewidth]{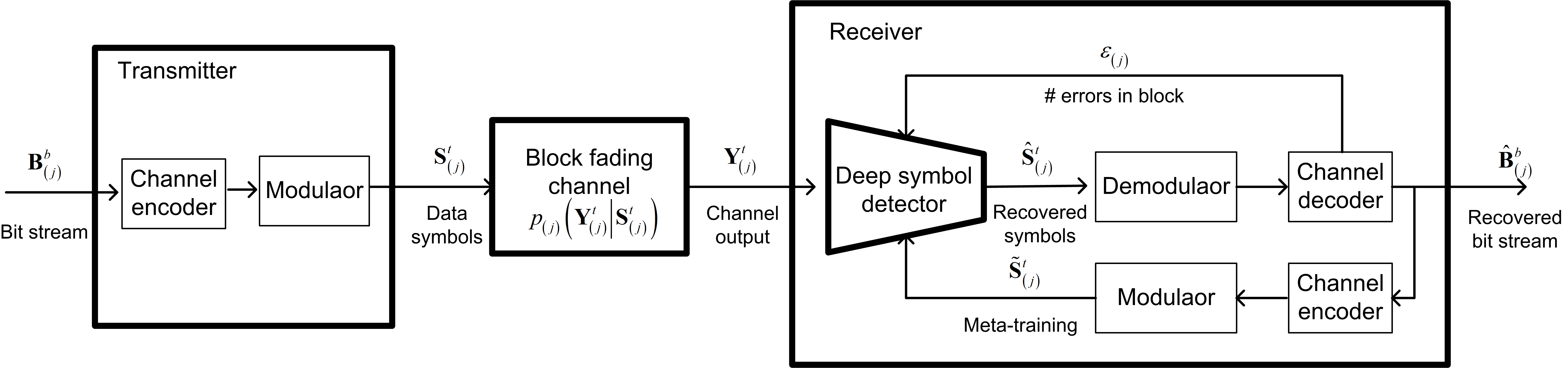} 
	\caption{Online training model.} 
	\label{fig:FeedbackModel2}
\end{figure*}	

ViterbiNet uses the channel outputs over a single block, denoted $\myVec{Y}^\Blklen\Blkidx$, where $j \in \mySet{N}$ represents the block index, to generate an estimate of the transmitted symbols, denoted $\hat{\myVec{S}}^\Blklen\Blkidx$.
In the presence of \ac{fec} codes, when the number of detection errors is not larger than the minimal distance of the code, the encoded bits can still be perfectly recovered \cite[Ch. 8]{Goldsmith:05}. It therefore follows that when the \ac{ser} of ViterbiNet is small enough such that the \ac{fec} code is able to compensate for these errors, the transmitted message can be recovered. The recovered message may also be re-encoded to generate new training samples, denoted $\tilde{\myVec{S}}^\Blklen\Blkidx$, referred to henceforth as {\em meta-training}. If indeed the encoded bits were successfully recovered, then $\tilde{\myVec{S}}^\Blklen\Blkidx$ represents the true channel input from which $\myVec{Y}^\Blklen\Blkidx$ was obtained. Consequently, the pair $\tilde{\myVec{S}}^\Blklen\Blkidx$ and $\myVec{Y}^\Blklen\Blkidx$ can be used to re-train ViterbiNet. The simple \ac{dnn} structure used in ViterbiNet implies that it can be efficiently and quickly retrained with a relatively small number of training samples, such as the size of a single transmission block in a block-fading channel. An illustration of the proposed online training mechanism is depicted in Fig. \ref{fig:FeedbackModel2}. 

A natural drawback of decision-directed approaches is their sensitivity to decision errors. For example, if the \ac{fec} decoder fails to successfully recover the encoded bits, then the meta-training samples $\tilde{\myVec{S}}^\Blklen\Blkidx$ do not accurately represent the channel input resulting in $\myVec{Y}^\Blklen\Blkidx$. In such cases, the inaccurate training sequence may gradually deteriorate the recovery accuracy of the re-trained ViterbiNet, making the proposed approach unreliable, particularly in low \acp{snr} where decoding errors occur frequently. 
Nonetheless, when error detection codes are present in addition to \ac{fec} codes, the effect of decision errors can be mitigated. In particular, when the receiver has a reliable estimate of $\epsilon\Blkidx$, namely,  the number of bit errors in decoding the $j$th block, it can decide to generate meta-training and re-train the symbol detector only when the number of errors is smaller than some threshold  $\bar{\epsilon}$. Using this approach, only accurate meta-training is used, and the effect of decision errors can be mitigated and controlled. The proposed online training mechanism is summarized below as Algorithm \ref{alg:Algo2}. 
\begin{algorithm}
	\caption{Online Training of ViterbiNet}
	\label{alg:Algo2}
	\begin{algorithmic}[1]
		\STATE \underline{Input}: Block of channel outputs $\myVec{y}^{ \Blklen}$; ViterbiNet \ac{dnn} weights $\myVec{\theta}$ and mixture model parameters $\myVec{\varphi}$; Bit error threshold $\bar{\epsilon}$.
		\STATE \label{stp:OT1} Apply ViterbiNet with parameters $\myVec{\theta}, \myVec{\varphi}$ to recover the symbols $\hat{\myVec{s}}^{ \Blklen}$ from  $\myVec{y}^{ \Blklen}$.
		\STATE Decode bits vector $\hat{\myVec{b}}^\Codlen$	and number of error $\epsilon$ from $\hat{\myVec{s}}^{ \Blklen}$.
		\IF{ $\epsilon < \bar{\epsilon}$}	
			\STATE Encode and modulate $\hat{\myVec{b}}^\Codlen$ into $\tilde{\myVec{s}}^{ \Blklen}$. 
			\STATE\label{stp:OT2a} Re-train \ac{dnn}  weights $\myVec{\theta}$ using meta-training $\left( \tilde{\myVec{s}}^{ \Blklen}, \myVec{y}^{ \Blklen}\right) $. 
			\STATE\label{stp:OT2b} Update mixture model $\myVec{\varphi}$ using $ \myVec{y}^{ \Blklen}$.
		\ENDIF
		\STATE  \underline{Output}: Decoded bits $\hat{\myVec{b}}^\Codlen$;  ViterbiNet \ac{dnn} weights $\myVec{\theta}$ and mixture model parameters $\myVec{\varphi}$.
	\end{algorithmic}
\end{algorithm} 

\revision{Setting the value of $\bar{\epsilon}$ should be based on the nature of the expected variations between blocks as well as on the \ac{fec} scheme. When operating with gradual variations, one can set $\bar{\epsilon}$ to represent zero errors, guaranteeing that the meta-training symbols used for re-training ViterbiNet are accurate. However, this setting may limit the ability of ViterbiNet to track moderate channel variations. Using larger values of $\bar{\epsilon}$  should also account for the \ac{fec} method used, as for some codes, encoding a bit stream with a small number of errors may result in a distant codeword compared to the one transmitted, thus yielding inaccurate meta-training labels. In Subsection \ref{subsec:Sim_BF} we numerically evaluate the effect of different values of  $\bar{\epsilon}$  on the ability of ViterbiNet to reliably track block-fading channels. }

\label{txt:RetrainRate}
\revision{
The learning rate used in re-training the \ac{dnn} in Step \ref{stp:OT2a} of Algorithm \ref{alg:Algo2} should be set based on the expected magnitude of the channel variations. Specifically, the value of the learning rate balances the contributions of the initial \ac{dnn} weights, which were tuned based on a past training set corresponding to previous channel realizations, and that of the typically much smaller meta-training set, representing the current channel. Using larger learning rates implies that the network can adapt to nontrivial channel variations, but also increases the sensitivity of the algorithm to decision errors, and should thus be used with smaller values of the threshold $\bar{\epsilon}$.}

The method for updating the mixture model parameters $\myVec{\varphi}$ in Step \ref{stp:OT2b} of Algorithm \ref{alg:Algo2} depends on which of the possible finite mixture model estimators discussed in Subsection \ref{subsec:Derivation} is used. For example, when using \ac{em}-based model fitting, the previous model parameters  $\myVec{\varphi}$ can be used as the initial guess used for fitting the \ac{pdf} based on the current channel outputs.

Finally, we note that our meta-learning based online training scheme exploits only the structure induced by channel codes. Modern communication protocols dictate a multitude of additional structures that can be exploited to generate meta-training.  
For example, many communication protocols use a-priori known preamble sequences to facilitate synchronization between the communicating nodes \cite[Ch. 14.2]{Goldsmith:05}. The receiver can thus use the a-priori knowledge of the preamble to generate meta-training from the corresponding channel outputs, which can be used for re-training ViterbiNet online. Additional structures induced by protocols that can be exploited to re-train ViterbiNet in a similar manner include specific header formats, periodic pilots, and network management frames.   
Consequently, while the proposed approach focuses on using channel codes to generate meta-training, it can be extended to utilize various other inherent structures of digital communication signals. 

	
	\vspace{-0.2cm}
	\section{Numerical Study}
	\label{sec:Sims}
	\vspace{-0.1cm}
	In this section we numerically evaluate the performance of ViterbiNet.  
	In particular, we first consider time-invariant channels in Subsection \ref{subsec:Sim_TI}. For such setups, we numerically compare the performance of ViterbiNet to the conventional model-based Viterbi algorithm as well as to previously proposed deep symbol detectors, and evaluate the resiliency of ViterbiNet to inaccurate training. Then, we consider block-fading channels, and evaluate the performance of ViterbiNet with online training in Subsection \ref{subsec:Sim_BF}. 
	In the following we assume that the channel memory $\Mem$ is a-priori known. It is emphasized that for all the simulated setups we were able to accurately detect the memory length using a standard correlation peak based estimator\footnote{We note that correlation based channel length estimators assume that the transmitter and the receiver are synchronized in time.}, see, e.g., \cite{Nguyen:07}, hence this assumption is not essential.
	\label{txt:Trainingdetails}
	Throughout this numerical study we implement the fully connected network in Fig. \ref{fig:NetworkArchitecture} using three layers: a $1 \times 100$ layer followed by a $100 \times 50$ layer and a $50 \times  16 (=|\CnstSize|^{\Mem})$ layer, using intermediate sigmoid and ReLU activation functions, respectively. The mixture model estimator approximates the distribution as a Gaussian mixture using \ac{em}-based fitting \cite[Ch. 2]{McLachlan:04}. \revision{The network is trained using $5000$ training samples to minimize the cross-entropy loss via the Adam optimizer \cite{Kingma:14} with learning rate $0.01$, using up to $100$ epochs with mini-batch size of $27$ observations.} We note that the number of training samples is of the same order and even smaller compared to typical preamble sequences in wireless communications\footnote{For example, preamble sequences in fourth-generation cellular LTE systems consist of up to $10$ subframes of $2.4\cdot 10^{4}$ samples, each embedding $839$ symbols \cite[Ch. 17]{Dahlman:10}.}. Due to the small number of training samples and the simple architecture of the \ac{dnn}, only a few minutes are required to train the network on a standard CPU.

\vspace{-0.2cm}
\subsection{Time-Invariant Channels}
\label{subsec:Sim_TI}
\vspace{-0.1cm}
	 We first consider two finite-memory causal scalar channels: An \ac{isi} channel with \ac{awgn}, and a Poisson channel. In both channels we set the memory length to $\Mem = 4$. 
	 For the \ac{isi} channel with \ac{awgn}, we let $W[i]$ be a zero-mean unit variance \ac{awgn} independent of $S[i]$, and let $\myVec{h} (\gamma)\in \mySet{R}^\Mem$ be the channel vector representing an exponentially decaying profile given by $\left( \myVec{h}\right)_\tau \triangleq e^{-\gamma(\tau-1)}$ for $\gamma > 0$. The  input-output relationship is given by
	 \begin{equation}
	 \label{eqn:AWGNCh1}
	 Y[i] = \sqrt{\rho} \cdot\sum\limits_{\tau=1}^{\Mem} \left( \myVec{h}(\gamma)\right)_\tau S[i-\tau + 1] + W[i],
	 \end{equation}
	 where $\rho > 0$ represents the \ac{snr}. Note that even though the effect of the \ac{isi} becomes less significant as $\gamma$ grows,  the channel \eqref{eqn:AWGNCh1} has memory length of $\Mem = 4$ for all values of $\gamma$ considered in this study.
	 The channel input is randomized from a binary phase shift keying constellation, i.e., $\mySet{S} = \{-1, 1\}$. 
	 For the Poisson channel, the channel input represents on-off keying, namely, $\mySet{S} = \{0,1\}$, and the channel output $Y[i]$ is generated from the input via 
	 \begin{equation}
	\label{eqn:PoissonCh1}
	Y[i] | \myVec{S}^\Blklen\sim \mathds{P}\left( \sqrt{\rho} \cdot\sum\limits_{\tau=1}^{\Mem} \left( \myVec{h}(\gamma)\right)_\tau S[i-\tau + 1] + 1\right),
	 \end{equation}
	 where $\mathds{P}(\lambda)$ is the Poisson distribution with parameter $\lambda > 0$.
	
	\label{txt:SNRTrain}
 	\revision{For each channel, we numerically compute the \ac{ser} of  ViterbiNet for  different values of the \ac{snr} parameter $\rho$. In the following study, the \ac{dnn} is trained anew for each value of $\rho$.}
	For each \ac{snr} $\rho$, the \ac{ser} values are averaged over $20$ different channel vectors  $\myVec{h} (\gamma)$, obtained by letting $\gamma$ vary in the range $[0.1, 2]$. 
	 For comparison, we  numerically compute the \ac{ser} of the Viterbi algorithm, as well as that of the \ac{sbrnn} deep symbol decoder proposed in \cite{Farsad:18}. 
	 \label{txt:Robustness1}
	 In order to study the resiliency of ViterbiNet to inaccurate training, we also compute the performance when the receiver only has access to a noisy estimate of $\myVec{h}(\gamma)$, and specifically, to a copy of $\myVec{h}(\gamma)$ whose entries are corrupted by i.i.d. zero-mean Gaussian noise with variance $\sigma_e^2$. In particular, we use  $\sigma_e^2 = 0.1$ for the Gaussian channel \eqref{eqn:AWGNCh1}, and $\sigma_e^2 = 0.08$ for the Poisson channel \eqref{eqn:PoissonCh1}. 
	\revision{We consider two cases: {\em Perfect \ac{csi}}, in which the channel-model-based Viterbi detector has accurate knowledge of  $\myVec{h}(\gamma)$, while ViterbiNet is trained using labeled samples generated with the same  $\myVec{h}(\gamma)$ used for generating the test data; and {\em \ac{csi} uncertainty}, where the Viterbi algorithm implements Algorithm \ref{alg:Algo1} with the log-likelihoods computed using the noisy version of  $\myVec{h}(\gamma)$,  while the labeled data used for training ViterbiNet is generated with  the noisy version of $\myVec{h}(\gamma)$ instead of the true one. In particular, for \ac{csi} uncertainty, the $5000$ samples used for training ViterbiNet are divided into $10$ subsets, each generated from a channel with a different realization of the noise in $\myVec{h}(\gamma)$. In all cases, the information symbols are uniformly randomized in an i.i.d. fashion from $\mySet{S}$, and the test samples are generated from their corresponding channel, i.e., \eqref{eqn:AWGNCh1} and \eqref{eqn:PoissonCh1} for the Gaussian and Poisson channels, respectively, with the true  channel vector $\myVec{h}(\gamma)$.} 
%
	
\begin{figure}
			\centering
			\includegraphics[ width=\figWidth]{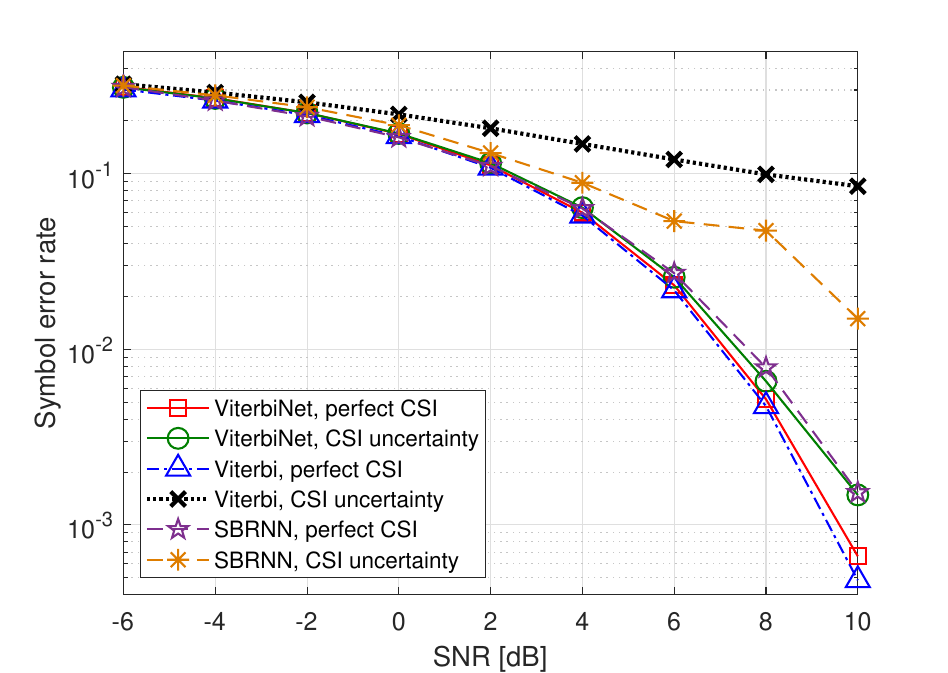}
			\figSpace
			\caption{\ac{ser} versus \ac{snr}, \ac{isi} channel with \ac{awgn}.
			}
			\label{fig:AWGN}
		\end{figure}

\begin{figure}
			\centering
			\includegraphics[ width=\figWidth]{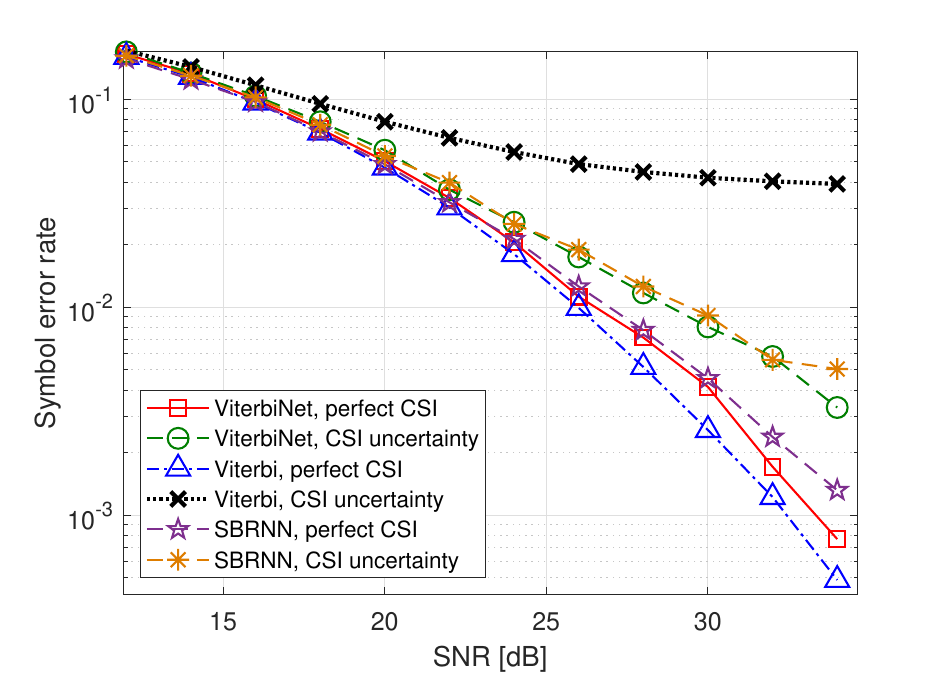}
			\figSpace
			\caption{\ac{ser} versus \ac{snr}, Poisson channel.
			}
			\label{fig:Poisson}
		\end{figure}

	The numerically computed \ac{ser} values, averaged over $50000$ Monte Carlo simulations, versus $\rho \in [-6,10]$ dB for the \ac{isi} channel with \ac{awgn} are depicted in Fig. \ref{fig:AWGN}, while the corresponding performance versus $\rho \in [10,34]$ dB for the Poisson channel are depicted in Fig. \ref{fig:Poisson}. Observing Figs. \ref{fig:AWGN}-\ref{fig:Poisson}, we note that the performance of ViterbiNet approaches that of the conventional \ac{csi}-based Viterbi algorithm. 
	In particular, for the \ac{awgn} case, in which the channel output obeys a Gaussian mixture distribution, the performance of ViterbiNet coincides with that of the Viterbi algorithm, while for the Poisson channel a very small gap is observed at high \acp{snr} due to the model mismatch induced by approximating the distribution of $Y[i]$ as a Gaussian mixture. We expect the performance of ViterbiNet to improve in that scenario when using a Poisson kernel density estimator for $\PdfEst{\myVec{\varphi}}\left(  y[i]\right)$, however, this requires some basic level of knowledge of the input-output relationship.  
	It is also observed in  Figs. \ref{fig:AWGN}-\ref{fig:Poisson} that the \ac{sbrnn} receiver, which was shown in \cite{Farsad:18} to approach the performance of the \ac{csi}-based Viterbi algorithm when sufficient training is provided, is outperformed by ViterbiNet here due to the small amount of training data provided. 
	For example, for the \ac{isi} channel with \ac{awgn}, both the Viterbi detector as well as ViterbiNet achieve an \ac{ser} of  $4.7 \cdot 10^{-3}$ for \ac{snr} of $8$ dB, while the \ac{sbrnn} detector achieves an \ac{ser} of $8.5 \cdot 10^{-3}$ for the same \ac{snr} value. For the Poisson channel at an \ac{snr} of $28$ dB, the Viterbi algorithm achieves an \ac{ser} of $5.1 \cdot 10^{-3}$, while ViterbiNet and the \ac{sbrnn} detector achieve \ac{ser} values of $6.6 \cdot 10^{-3}$ and $8.3 \cdot 10^{-3}$, respectively.
	These results demonstrate that ViterbiNet, which uses simple \ac{dnn} structures embedded into the Viterbi algorithm, requires significantly less training compared to previously proposed \ac{ml}-based receivers.
	
	In the presence of \ac{csi} uncertainty, it is observed in Figs. \ref{fig:AWGN}-\ref{fig:Poisson} that ViterbiNet significantly outperforms the Viterbi algorithm. In particular, when ViterbiNet is trained with a variety of different channel conditions, it is still capable of achieving relatively good \ac{ser} performance under each of the channel conditions for which it is trained, while the performance of the conventional Viterbi algorithm is significantly degraded in the presence of imperfect \ac{csi}. While the \ac{sbrnn} receiver is shown to be more resilient to inaccurate \ac{csi} compared to the Viterbi algorithm, as was also observed in \cite{Farsad:18}, it is outperformed by ViterbiNet with the same level of uncertainty, and the performance gap is more notable in the \ac{awgn} channel. The reduced gain of ViterbiNet over the \ac{sbrnn} receiver for the Poisson channel stems from the fact that ViterbiNet here uses a Gaussian mixture kernel density estimator for the \ac{pdf} of $Y[i]$, which obeys a Poisson mixture distribution for the Poisson channel \eqref{eqn:PoissonCh1}. 
	

\begin{figure}
	\centering
	\includegraphics[ width=\figWidth]{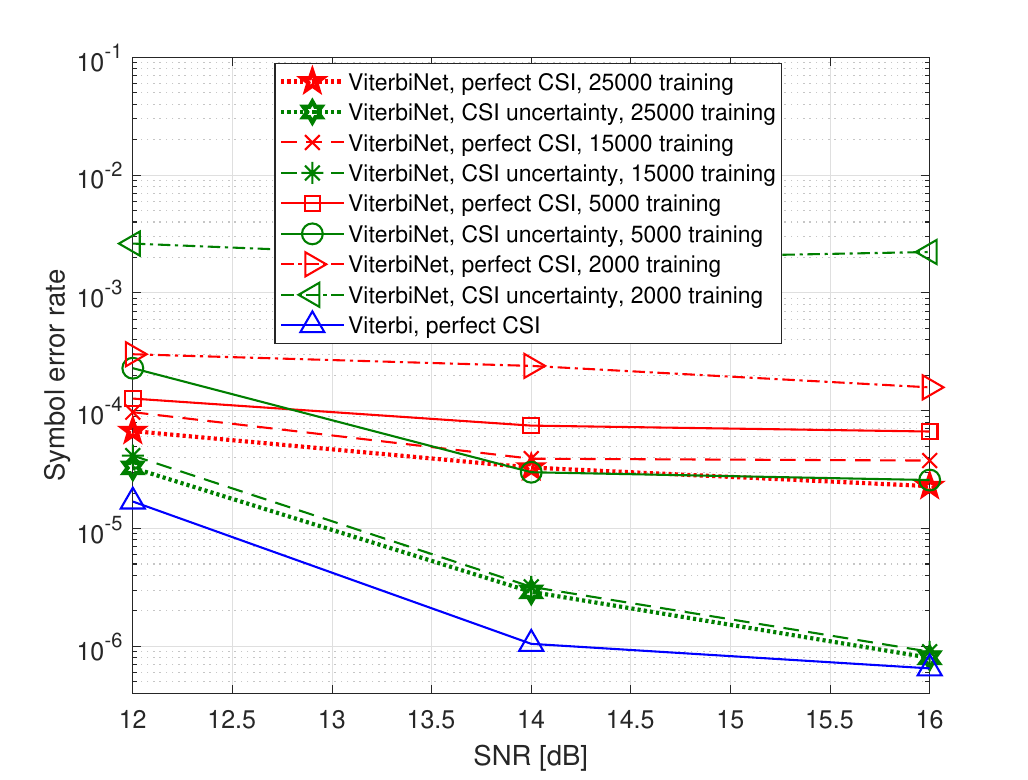}
	\figSpace
	\caption{High-\ac{snr} performance for different training sizes, \ac{isi} channel with \ac{awgn}.
	}
	\label{fig:HighSNR}
\end{figure}

\revision{
The simulation results depicted in Figs. \ref{fig:AWGN}-\ref{fig:Poisson} focus on \acp{snr}  for which the \ac{ser} values are of the orders of $10^{-2}$ and $10^{-3}$. 
While it is observed in Figs. \ref{fig:AWGN}-\ref{fig:Poisson} that ViterbiNet approaches the performance of the Viterbi algorithm with perfect \ac{csi}, it is also noted that there is a small \ac{ser} gap between these symbol detector in higher \acp{snr}. This gap indicates that training is more difficult in higher \acp{snr}, for which there is a smaller diversity in the training samples compared to lower \ac{snr} values, and thus larger training sets are required to improve performance. To demonstrate this, we depict in Fig. \ref{fig:HighSNR} the \ac{ser} values achieved by ViterbiNet when trained using training sets of sizes $\{2, 5, 15, 25\}\cdot 10^{3}$ for the \ac{isi} channel with \ac{awgn} with \acp{snr} of $12-16$ dB, i.e., higher \acp{snr} than those depicted in Fig. \ref{fig:AWGN}. The numerically evaluated performance, averaged here over $10^6$ Monte Carlo simulations to faithfully capture \ac{ser} values in the order of $10^{-5}$, is compared to the channel-model-based Viterbi algorithm with perfect \ac{csi}.  Observing Fig. \ref{fig:HighSNR}. we note that in high \acp{snr}, ViterbiNet requires larger training sets in order to approach the performance of the Viterbi algorithm, when trained using samples from the same channel under which it is tested, i.e., perfect \ac{csi}. However, it is also noted that, except for very small training sets, ViterbiNet with \ac{csi} uncertainty outperforms ViterbiNet with perfect \ac{csi}, and that its performance is within a small gap from that of the Viterbi detector. These results indicate that, in high \acp{snr}, the additional  diversity induced when training under \ac{csi} uncertainty is in fact beneficial, and allows ViterbiNet to achieve smaller \ac{ser} values compared to training using samples from the same channel realization.}

\begin{figure}
			\centering
			\includegraphics[ width=\figWidth]{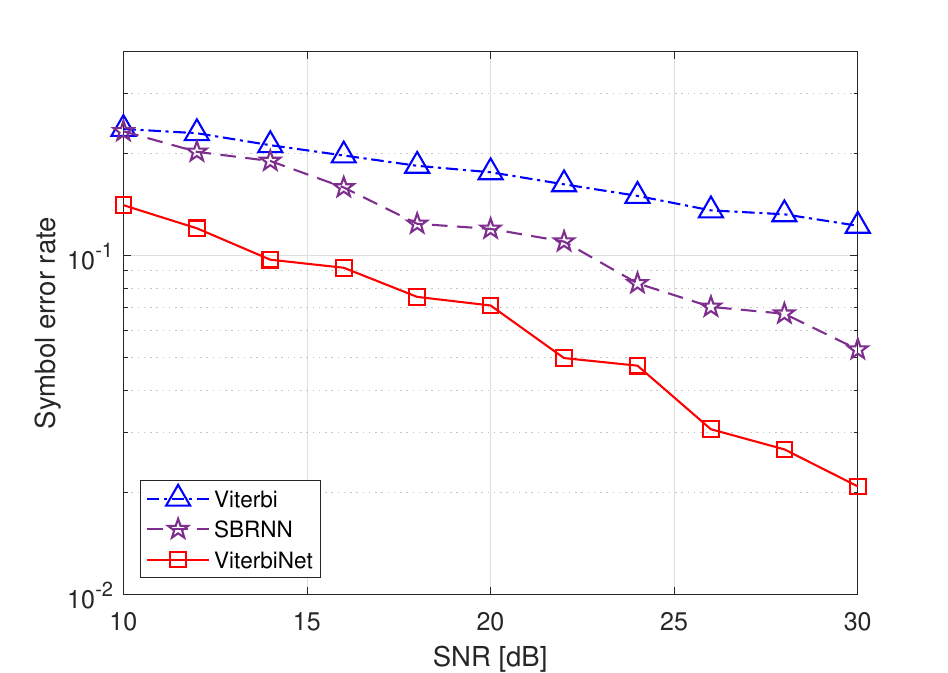}
			\figSpace
			\caption{\ac{ser} versus \ac{snr}, \ac{isi} channel with alpha-stable noise.
			}
			\label{fig:Alpha}
		\end{figure}

The main benefit of ViterbiNet is its ability to accurately implement the Viterbi algorithm in causal finite-memory channels while requiring only knowledge of the memory length, i.e., no \ac{csi} is required. 
The simulation study presented in Figs. \ref{fig:AWGN}-\ref{fig:Poisson} demonstrates an additional gain of ViterbiNet over the Viterbi algorithm, which is its improved resiliency to inaccurate \ac{csi}.  
Another benefit of ViterbiNet is its ability to reliably operate in setups where, even when full instantaneous \ac{csi} is available, the Viterbi algorithm is extremely complicated to implement. To demonstrate this benefit, we again consider the \ac{isi} channel in \eqref{eqn:AWGNCh1} with an exponential decay parameter $\gamma = 0.2$, however, here we set the distribution of the additive noise signal $W[i]$ to an {\em alpha-stable} distribution \cite[Ch. 1]{Nolan:03}. 
Alpha-stable distributions, which are obtained as a sum of heavy tailed independent \acp{rv} via the generalized central limit theorem, are used to model the noise in various communications scenarios, including molecular channels \cite{Farsad:17}, underwater acoustic channels \cite{Pelekanakis:15}, and impulsive noise channels \cite{Johnston:17}. In particular, we simulate an alpha-stable noise  with stability parameter $\alpha = 0.5$, skewness parameter $\beta = 0.75$, scale parameter $c = 1$, and location parameter $\mu = 0$, following \cite[Ch. III]{Farsad:17}. 
Recall that the \ac{pdf} of alpha-stable \acp{rv} is in general not analytically expressible \cite[Ch. 1.4]{Nolan:03}. Thus, implementing the Viterbi algorithm for such channels is extremely difficult even when full \ac{csi} is available, as it requires accurate numerical approximations of the conditional \acp{pdf} to compute the log likelihoods. 
Consequently, in order to implement the Viterbi algorithm here, we numerically approximated the \ac{pdf} of such alpha-stable \acp{rv} using the direct integration method \cite[Ch. 3.3]{Nolan:03} over a grid of $50$ equally spaced points in the range $[-5,5]$, and used the \ac{pdf} of the closest grid point when computing the log-likelihood terms in the Viterbi detector. 
 In Fig. \ref{fig:Alpha} we depict the \ac{ser} performance of ViterbiNet compared to the \ac{sbrnn} detector and the Viterbi algorithm for this channel versus $\rho \in [10,30]$ dB. 
 Observing Fig. \ref{fig:Alpha}, we note that the \ac{csi}-based Viterbi detector, which has to use a numerical approximation of the \ac{pdf}, is notably outperformed by the data-driven symbol detectors, which are ignorant of the exact statistical model and do not require an analytical expression of the conditional \ac{pdf} to operate. Furthermore, we note that  ViterbiNet achieves a substantial \ac{snr} gap from the \ac{sbrnn} detector. For example, ViterbiNet achieves an \ac{ser} smaller than $5\cdot 10^{-2}$ for \ac{snr} values larger than $22$ dB, while the \ac{sbrnn} detector requires an \ac{snr} of at least $30$ dB to achieve a similar \ac{ser}, namely, an \ac{snr} gap of $8$ dB. Finally, the performance of ViterbiNet scales with respect to \ac{snr} in a similar manner as in Figs. \ref{fig:AWGN}-\ref{fig:Poisson}. 
 The results in Fig. \ref{fig:Alpha} demonstrate the ability of ViterbiNet to reliably operate under non-conventional channel models for which the Viterbi algorithm is difficult to accurately implement, even when full \ac{csi} is available.

\begin{figure}
	\centering
	\includegraphics[ width=\figWidth]{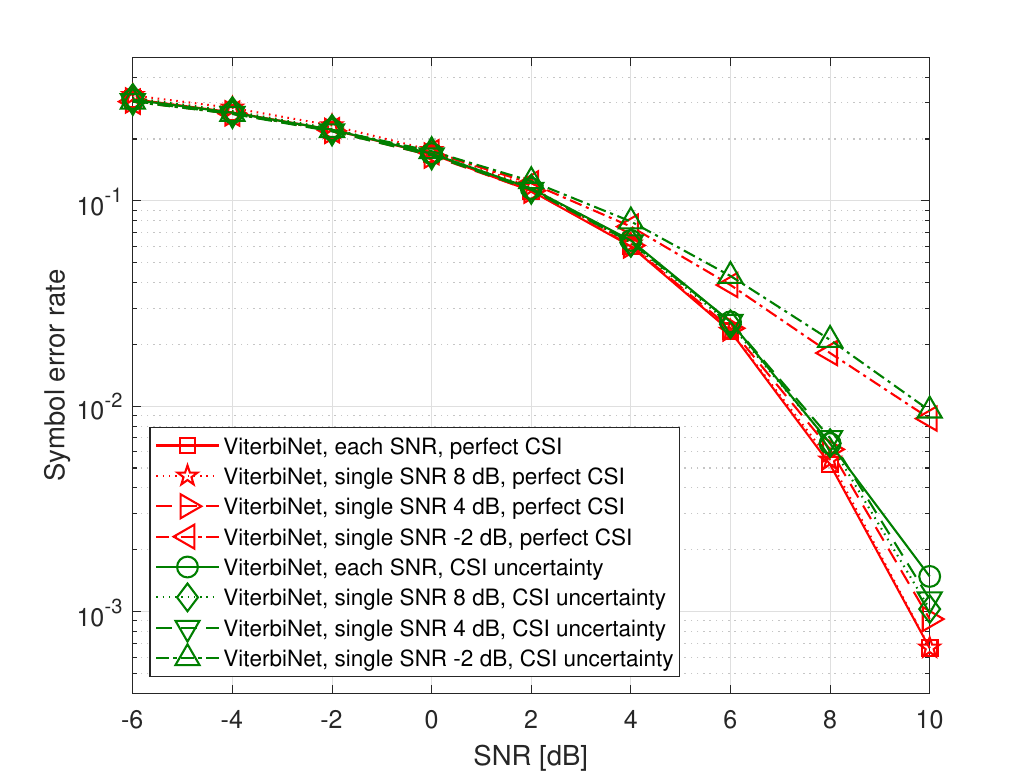}
	\figSpace
	\caption{Training for each \ac{snr} compared to training over a single \ac{snr}, \ac{isi} channel with \ac{awgn}.
	}
	\label{fig:SingleSNR}
\end{figure}

\revision{
Finally, we study the robustness of ViterbiNet to  different noise powers than those used during training. Recall that so far we have trained ViterbiNet for each value of $\rho$, i.e., for each \ac{snr}. To evaluate its resiliency to an inaccurate noise level in training, we next numerically compute the \ac{ser} of  ViterbiNet for  different values of the \ac{snr} parameter $\rho$. 
In particular, we repeat the simulation study of the \ac{isi} channel with \ac{awgn}, whose results are depicted in Fig. \ref{fig:AWGN}, while training ViterbiNet using $5000$ samples drawn from a channel with a single \ac{snr}, i.e., ViterbiNet is trained once instead of being re-trained for each \ac{snr}. The numerically computed \ac{ser} values are depicted in Fig. \ref{fig:SingleSNR}, where the curves denoted {\em 'ViterbiNet, single SNR'} represent the performance achieved when training only with samples corresponding to a single \ac{snr}, which is in the set $\{-2, 4, 8\}$ dB; The curves denoted {\em 'ViterbiNet, each SNR'} stand for the performance achieved when ViterbiNet is trained anew for each \ac{snr} level, i.e., the same curves as those of Fig. \ref{fig:AWGN} of the revised manuscript. 
Observing Fig. \ref{fig:SingleSNR}, we note that ViterbiNet achieves approximately the same performance when  trained using a single \ac{snr} value of either $4$ dB or $8$ dB as when it is trained anew for each \ac{snr}. 
A small gap is observed when ViterbiNet is trained using samples corresponding to a low \ac{snr} level of $-2$ dB.
These results demonstrate the ability of ViterbiNet to generalize to multiple \ac{snr} levels.}

\vspace{-0.2cm}
\subsection{Block-Fading Channels}
\label{subsec:Sim_BF}
\begin{figure}
			\centering
			\includegraphics[ width=\figWidth]{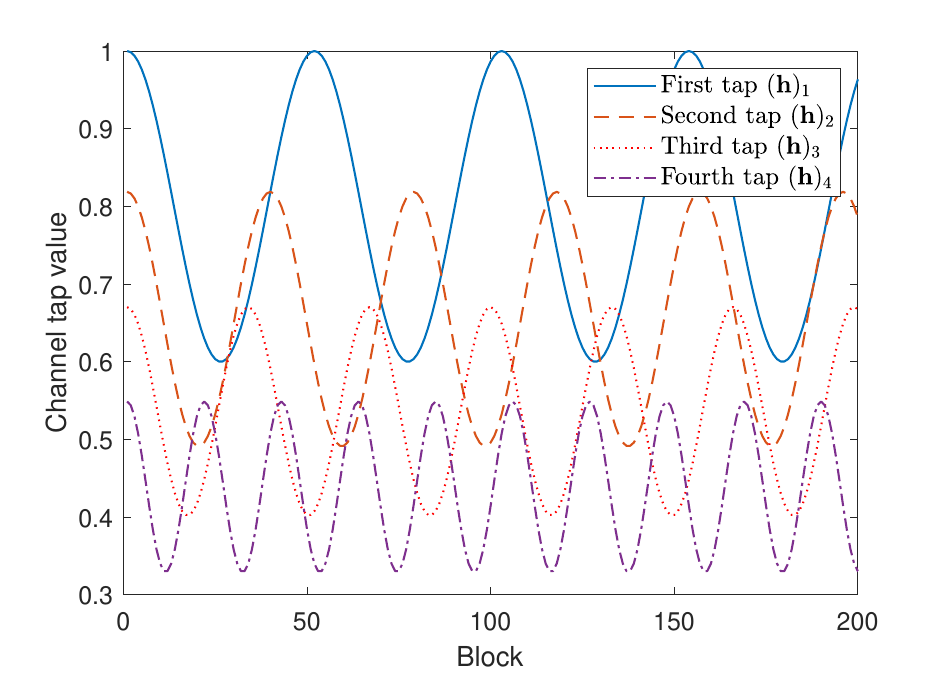}
			\figSpace
			\caption{Channel coefficient variations between different blocks.
			}
			\label{fig:TVChannel} 
		\end{figure}

\vspace{-0.1cm}
	Next, we numerically evaluate the online training method for tracking block-fading channels detailed in Section \ref{sec:BlockFading}. 
	To that aim, we consider two block-fading channels: an \ac{isi} channel with \ac{awgn} and a Poisson channel. For each transmitted block we use the channel models in \eqref{eqn:AWGNCh1} and \eqref{eqn:PoissonCh1}, respectively, with a fixed exponential decay parameter $\gamma = 0.2$, and thus we omit the notation $\gamma$ from the vector $\myVec{h}$ in this subsection. However, unlike the scenario studied in the previous subsection, here the channel coefficient vector $\myVec{h}$ varies between blocks. In particular, for the $j$th block we use $\big(\myVec{h}\Blkidx \big)_{\tau} \triangleq e^{-0.2(\tau-1)}\cdot\Big(0.8 + 0.2\cos\big( \frac{2\pi \cdot j}{\left( \myVec{p}\right)_\tau}\big)  \Big)$, with $ \myVec{p} = [51, 39, 33, 21]^T$. An illustration of the variations in the channel coefficients with respect to the block index $j$ is depicted in Fig. \ref{fig:TVChannel}.   	

\begin{figure}
			\centering
			\includegraphics[ width=\figWidth]{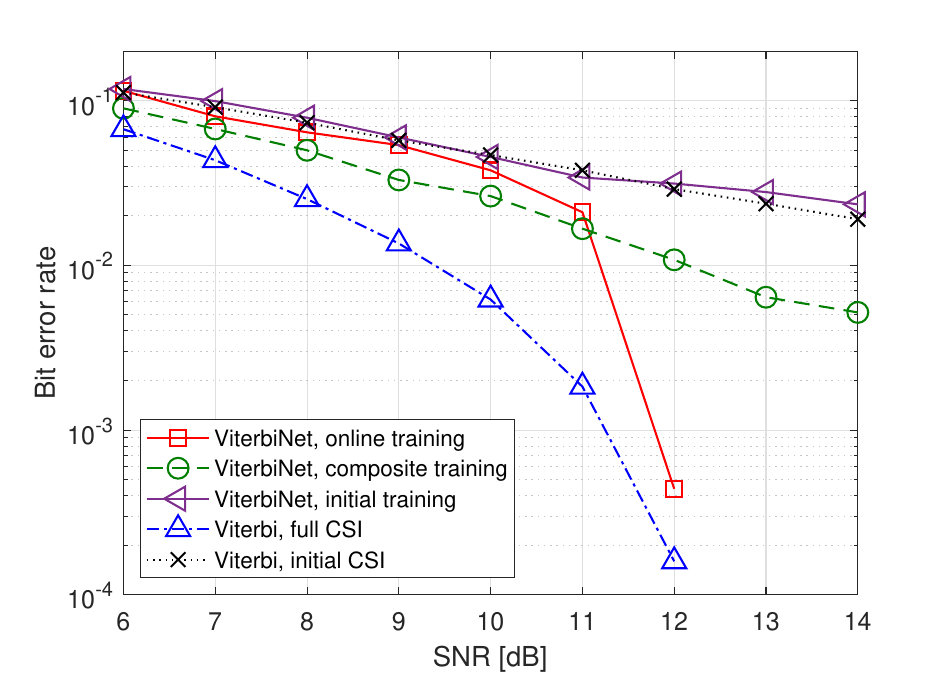}
			\figSpace
			\caption{Coded \ac{ber} versus \ac{snr}, block-fading \ac{isi} channel with \ac{awgn}.
			}
			\label{fig:TVChannelAWGN} 
		\end{figure}

\begin{figure}
			\centering
			\includegraphics[ width=\figWidth]{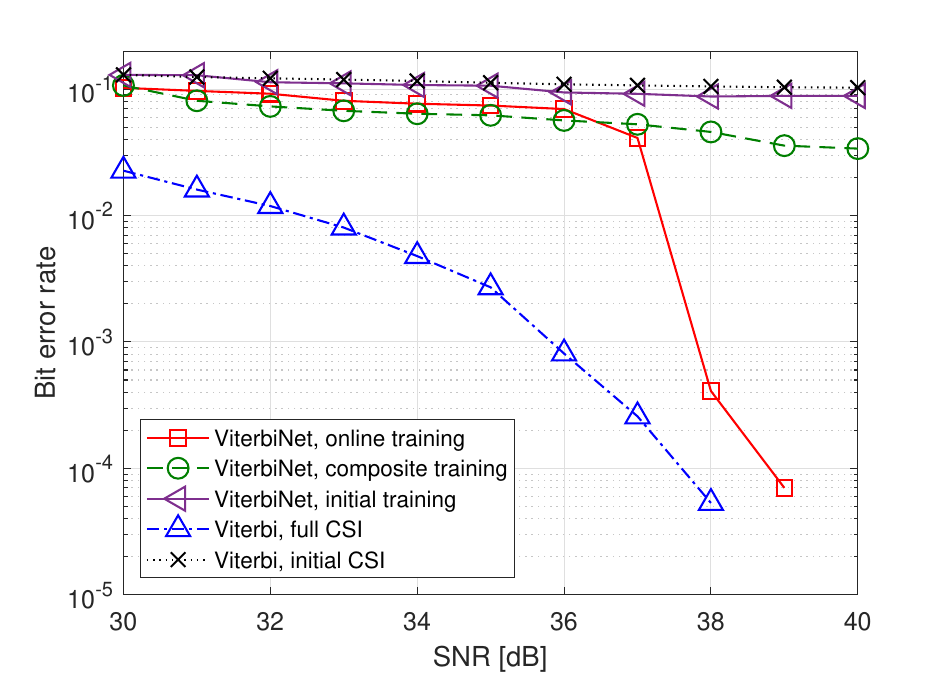}
			\figSpace
			\caption{Coded \ac{ber} versus \ac{snr}, block-fading Poisson channel.
			}
			\label{fig:TVChannelPoisson} 
		\end{figure}

	In each channel block a single codeword of an \ac{rs} [255, 223] \ac{fec} code is transmitted, namely, during each block $\Codlen = 1784$ bits are conveyed using $\Blklen = 2040$ binary symbols. 
	Before the first block is transmitted, ViterbiNet is trained using $5000$ training samples taken using the initial channel coefficients, i.e., using the initial channel coefficients vector $\myVec{h}_{(1)}$. 
	\label{txt:LearnRate2}
	\revision{For online training, we use a learning rate of $0.002$, i.e., five times smaller than that used for initial training, and a bit error threshold of $\bar{\epsilon} = 2 \%$.}
	In addition to ViterbiNet with online training, we also compute the coded \ac{ber} performance of ViterbiNet when trained only once using the $5000$ training samples representing the initial channel  $\myVec{h}_{(1)}$, referred to as {\em initial training}, as well as ViterbiNet trained once using $5000$ training samples corresponding to $10$ different channels $\{\myVec{h}_{(3\cdot k)}\}_{k=1}^{10}$, referred to as {\em composite training}. The performance of ViterbiNet is compared to the Viterbi algorithm with full instantaneous \ac{csi} as well as to the Viterbi detector which assumes that the channel is the time-invariant initial channel $\myVec{h}_{(1)}$. The coded \ac{ber} results, averaged over $200$ consecutive blocks, are depicted in Figs. \ref{fig:TVChannelAWGN}-\ref{fig:TVChannelPoisson} for the \ac{isi} channel with \ac{awgn} and for the Poisson channel, respectively. 
 	
	
	Observing Figs. \ref{fig:TVChannelAWGN}-\ref{fig:TVChannelPoisson}, we note that for both channels, as the \ac{snr} increases, ViterbiNet with online training approaches the performance of the \ac{csi}-based Viterbi algorithm which requires accurate knowledge of the complete input-output statistical relationship for each block, while for low \acp{snr}, its performance is only slightly improved compared to that of the Viterbi algorithm using only the initial channel. This can be explained by noting that for high \ac{snr} values, the \ac{ber} at each block is smaller than the threshold  $\bar{\epsilon}$, and thus our proposed online training scheme is capable of generating reliable meta-training, which ViterbiNet uses to accurately track the channel, thus approaching the optimal performance. However, for low \ac{snr} values, the \ac{ber} is typically larger than $\bar{\epsilon}$, thus ViterbiNet does not frequently update its coefficients in real-time, thus achieving  only a minor improvement over using only the initial  training data set. It is emphasized that increasing the \ac{ber} threshold  $\bar{\epsilon}$ can severely deteriorate the performance of ViterbiNet at low \acp{snr}, as it may be trained online using inaccurate meta-training.
	
	We also observe in Figs. \ref{fig:TVChannelAWGN}-\ref{fig:TVChannelPoisson} that the performance ViterbiNet with initial training coincides with that of Viterbi algorithm using the initial channel, as both detectors effectively implement the same detection rule, which is learned by ViterbiNet from the initial channel training set. Furthermore, the composite training approach allows ViterbiNet to achieve improved \ac{ber} performance compared to using only initial training, as the resulting decoder is capable of operating in a broader range of different channel conditions. Still, is is noted that for both the \ac{awgn} channel and the Poisson channel, ViterbiNet with composite training is notably outperformed by the optimal Viterbi detector with instantaneous \ac{csi}, whose \ac{ber} performance is approached only when using online training at high \acp{snr}. 
	
\begin{figure}
		\centering
		\includegraphics[ width=\figWidth]{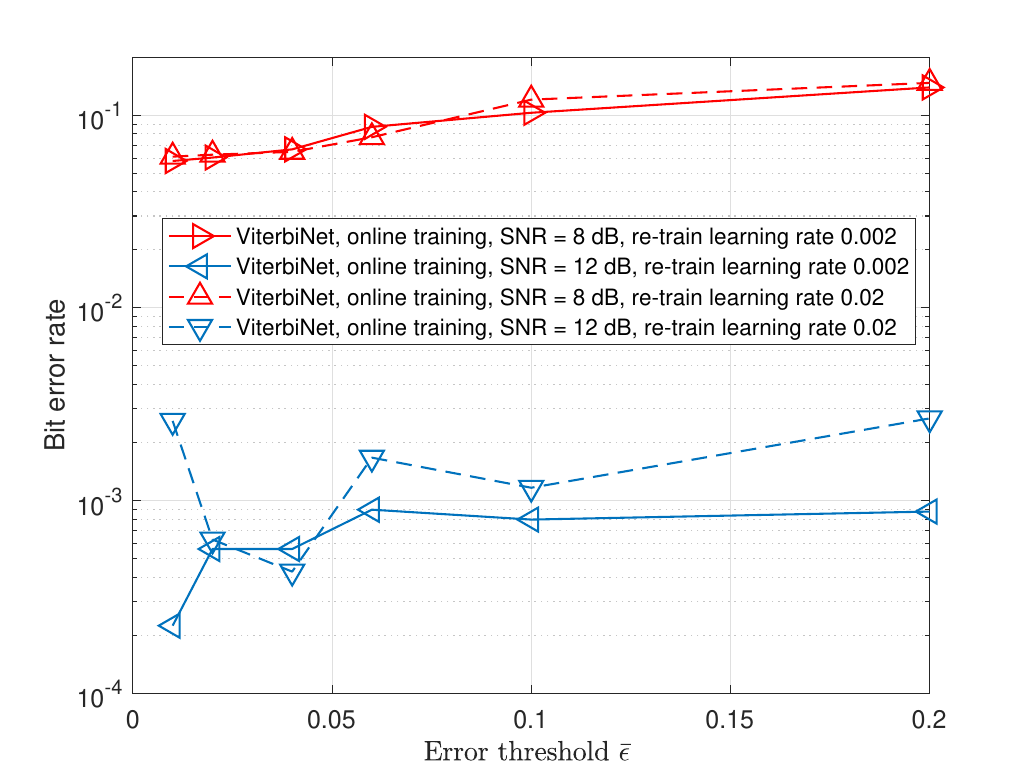}
		\figSpace
		\caption{ Coded \ac{ber} versus error threshold, block-fading \ac{isi} channel with \ac{awgn}.
		}
		\label{fig:CodedBER_Epsilon} 
	\end{figure}

	\revision{As discussed in Subsection \ref{subsec:Online}, the success of the meta-learning based method for training ViterbiNet online depends on the error threshold $\bar{\epsilon}$ and the learning rate used for re-training the \ac{dnn}. To numerically evaluate the effect of these parameters, we compare in Fig. \ref{fig:CodedBER_Epsilon} the \ac{ber} of ViterbiNet versus  $\bar{\epsilon}$ for the \ac{isi} with \ac{awgn} channel under \ac{snr} values of $8$ and $12$ dB, and with re-training learning rates of $0.002$ and $0.02$. These \ac{ber} values are compared to the  model-based Viterbi operating with full \ac{csi}. Observing Fig. \ref{fig:CodedBER_Epsilon} we note that, for the lower \ac{snr} of $8$ dB, a similar behavior, in which the \ac{ber} value increases with $\bar{\epsilon}$, is observed for both learning rates. However, for the higher \ac{snr} of $12$ dB, the \ac{ber} increase with the error threshold $\bar{\epsilon}$ for learning rate of $0.002$, while for the higher learning rate of $0.02$ we observe a minima around $\bar{\epsilon} = 0.04$. This behavior is because higher values of $\bar{\epsilon}$ may cause the network to use inaccurate training, while  values which are too small may result in the network being unable to reliably track channel variations.}
	
	The results presented in this section demonstrate that a data-driven symbol detector designed by integrating \acp{dnn} into the Viterbi algorithm is capable of approaching optimal performance without  \ac{csi}.   Unlike the \ac{csi}-based Viterbi algorithm, its \ac{ml}-based implementation demonstrates robustness to imperfect \ac{csi}, and achieves excellent performance in channels where the Viterbi algorithm is difficult to implement. 
	The fact that ViterbiNet requires a relatively small training set allows it to accurately track block-fading channels using meta-training generated from previous decisions on each transmission block. 
	We thus conclude that designing a data-driven symbol detector by combining \ac{ml}-based methods with the optimal model-based algorithm yields a reliable, efficient, and resilient data-driven symbol recovery mechanism.

	\vspace{-0.25cm}
	\section{Conclusions}
	\label{sec:Conclusions}
	\vspace{-0.15cm}
	We proposed an \ac{ml}-based implementation of the Viterbi symbol detection algorithm called ViterbiNet. 
	To design ViterbiNet, we identified the log-likelihood computation as the part of the Viterbi algorithm that requires full knowledge of the underlying channel input-output statistical relationship. We then integrated an \ac{ml}-based architecture designed to compute the log-likelihoods into the algorithm. The resulting architecture combines \ac{ml}-based processing with the conventional Viterbi symbol detection scheme. In addition, we proposed a meta-learning based method that allows ViterbiNet to track time-varying channels online. 
	Our numerical results demonstrate that ViterbiNet approaches the optimal performance of the \ac{csi}-based Viterbi algorithm and  outperforms previously proposed \ac{ml}-based symbol detectors using a small amount of training data. It is also illustrated that ViterbiNet is capable of reliably operating in the presence of \ac{csi} uncertainty, as well as in complex channel models where the Viterbi algorithm is extremely difficult to implement. Finally, it is shown that ViterbiNet can adapt via meta-learning in block-fading channel conditions. 
	

\end{document}

%% file: Preamble.tex
\definecolor{NewColor}{rgb}{0,0,0} 

\newcommand{\myVec}[1]{{\boldsymbol{#1}}}

\newcommand{\mySet}[1]{\mathcal{#1}}

\newcommand{\Pdf}[1]{p_{ { #1}} }
\newcommand{\PdfEst}[1]{\hat{p}_{ { #1}} }
\newcommand{\Mem}{l}			 			
\newcommand{\Blklen}{t}			 			
\newcommand{\Codlen}{b}			 			
\newcommand{\Blkset}{\mySet{T}}
\newcommand{\CnstSize}{m}			 			
\newcommand{\Blkidx}{_{(j)}}

\newcommand{\Vecdim}[1]{} 

\acrodef{adc}[ADC]{analog-to-digital convertor}
\acrodef{cs}[CS]{compressed sensing}
\acrodef{dtft}[DTFT]{discrete-time Fourier transform}
\acrodef{dnn}[DNN]{deep neural network}
\acrodef{ml}[ML]{machine learning}
\acrodef{rnn}[RNN]{recurrent neural network}
\acrodef{csi}[CSI]{channel state information}
\acrodef{map}[MAP]{maximum a-posteriori probability}
\acrodef{snr}[SNR]{signal-to-noise ratio}
\acrodef{bs}[BS]{base station} 
\acrodef{iot}[IOT]{Interent of Things}
\acrodef{mimo}[MIMO]{multiple-input multiple-output}
\acrodef{mse}[MSE]{mean-squared error}
\acrodef{pdf}[PDF]{probability density function}
\acrodef{rv}[RV]{random variable}
\acrodef{fec}[FEC]{forward error correction}
\acrodef{rs}[RS]{Reed-Solomon}
\acrodef{lti}[LTI]{linear time-invariant}
\acrodef{wss}[WSS]{wide-sense stationary}
\acrodef{psd}[PSD]{power spectral density}
\acrodef{ser}[SER]{symbol error rate} 
\acrodef{ber}[BER]{bit error rate} 
\acrodef{isi}[ISI]{intersymbol interference} 
\acrodef{lstm}[LSTM]{long short-term memory} 
\acrodef{em}[EM]{expectation minimization} 
\acrodef{awgn}[AWGN]{additive white Gaussian noise}
\acrodef{sbrnn}[SBRNN]{sliding bidirectional \ac{rnn}}